\documentclass{article}

    \PassOptionsToPackage{numbers, compress}{natbib}

\usepackage[preprint]{neurips_2024}

\usepackage[utf8]{inputenc} 
\usepackage[T1]{fontenc}    
\usepackage{hyperref}       
\usepackage{url}            
\usepackage{booktabs}       
\usepackage{amsfonts}       
\usepackage{nicefrac}       
\usepackage{microtype}      
\usepackage{xcolor}         
\usepackage{makecell}

\usepackage{graphicx}
\usepackage{amsmath}
\usepackage{mathtools}
\usepackage{multirow}
\usepackage{colortbl}
\usepackage{wrapfig}

\newcommand{\ra}[1]{\renewcommand{\arraystretch}{#1}}

\title{VMamba: Visual State Space Model}

\author{
    Yue~Liu \textsuperscript{1}
    \quad Yunjie~Tian\textsuperscript{1} 
    \quad Yuzhong~Zhao\textsuperscript{1}
    \quad Hongtian~Yu\textsuperscript{1} \\
    \textbf{\quad Lingxi~Xie\textsuperscript{2}
    \quad Yaowei~Wang\textsuperscript{3}
    \quad Qixiang~Ye\textsuperscript{1}
    \quad Jianbin~Jiao\textsuperscript{1}
    \quad Yunfan~Liu\textsuperscript{1\ddag}} \\
    {\textsuperscript{1} UCAS} \quad {\textsuperscript{2} Huawei Inc.} \quad {\textsuperscript{3} Pengcheng Lab.} \\
    \small{\texttt{\{liuyue171,tianyunjie19,zhaoyuzhong20,yuhongtian17\}@mails.ucas.ac.cn}}\\
    \small{\texttt{198808xc@gmail.com, wangyw@pcl.ac.cn}}, \small{\texttt{\{qxye,jiaojb,liuyunfan\}@ucas.ac.cn}}\\
}

\begin{document}

\maketitle

\begin{abstract}

Designing computationally efficient network architectures remains an ongoing necessity in computer vision.
In this paper, we adapt Mamba, a state-space language model, into VMamba, a vision backbone with linear time complexity.
At the core of VMamba is a stack of Visual State-Space (VSS) blocks with the 2D Selective Scan (SS2D) module. By traversing along four scanning routes, SS2D bridges the gap between the ordered nature of 1D selective scan and the non-sequential structure of 2D vision data, which facilitates the collection of contextual information from various sources and perspectives.
Based on the VSS blocks, we develop a family of VMamba architectures and accelerate them through a succession of architectural and implementation enhancements.
Extensive experiments demonstrate VMamba's promising performance across diverse visual perception tasks, highlighting its superior input scaling efficiency compared to existing benchmark models.
Source code is available at \url{https://github.com/MzeroMiko/VMamba}
\end{abstract}

\section{Introduction}\label{sec:introduction}

Visual representation learning remains as a fundamental research area in computer vision that has witnessed remarkable progress in the era of deep learning.
To represent complex patterns in vision data, two primary categories of backbone networks, \textit{i.e.}, Convolutional Neural Networks (CNNs)~\cite{vgg, Resnet2016, densenet, EfficientNet2019, liu2022convnet} and Vision Transformers (ViTs)~\cite{ViT2021, Swin2021, DeiT2021, hivit}, have been proposed and extensively utilized in a variety of visual tasks.
Compared to CNNs, ViTs generally demonstrate superior learning capabilities on large-scale data due to their integration of the self-attention mechanism~\cite{vaswani2017attention, ViT2021}.
However, the quadratic complexity of self-attention w.r.t. the number of tokens imposes substantial computational overhead in downstream tasks involving large spatial resolutions.

To address this challenge, significant efforts have been made to improve the efficiency of attention computation~\cite{tay2022efficient, Swin2021, dong2022cswin}.
However, existing approaches either restrict the size of the effective receptive field~\cite{Swin2021} or suffer from notable performance degradation across various tasks~\cite{katharopoulos2020transformers, wang2020linformer}.
This motivates us to develop a novel architecture for vision data, while maintaining the inherent advantages of the vanilla self-attention mechanism, \textit{i.e.}, global receptive fields and dynamic weighting parameters~\cite{han2021connection}.

\begin{figure}
    \begin{center}
        \includegraphics[width=\textwidth]{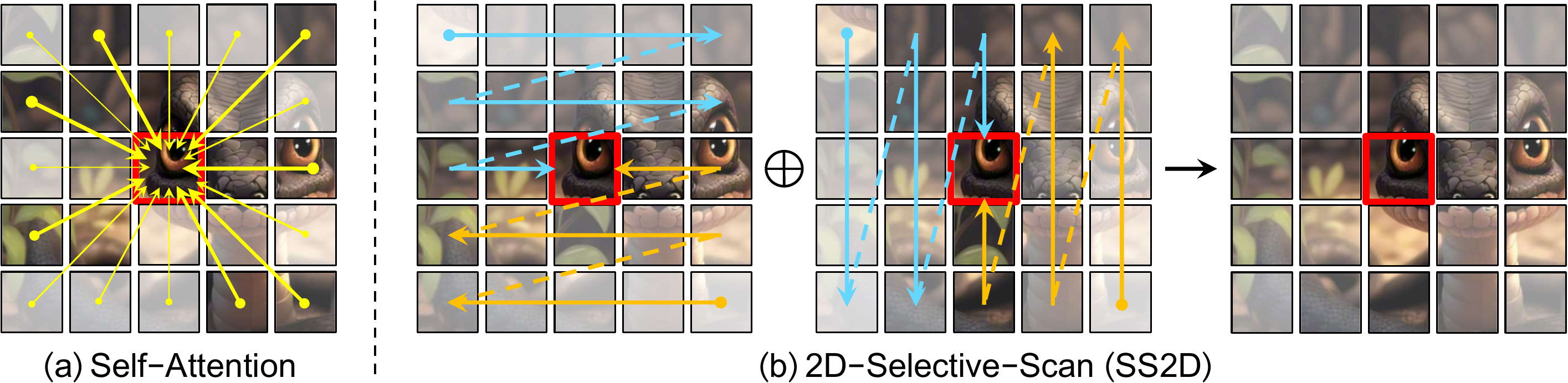}
        \caption{Comparison of the establishment of correlations between image patches through (a) self-attention and (b) the proposed 2D-Selective-Scan (SS2D). The red boxes indicate the query image patch, with its opacity representing the degree of information loss.
        } 
        \label{fig:att_comp}
    \end{center}
\end{figure}

Recently, Mamba~\cite{mambagu2023mamba}, a innovative State Space Model (SSM)~\cite{mambagu2023mamba, gssmehta2023long, selectives4wang2023selective, zubic2024state, schone2024scalable}, in the field of natural language processing (NLP), has emerged as a promising approach for long-sequence modeling with linear complexity.
Drawing inspiration from this advancement, we introduce VMamba, a vision backbone that integrates SSM-based blocks to enable efficient visual representation learning.
However, the core algorithm of Mamba, \textit{i.e.}, the parallelized selective scan operation, is essentially designed for processing one-dimensional sequential data. This presents a challenge when adapting it for processing vision data, which lacks an inherent sequential arrangement of visual components.
To address this issue, we propose 2D Selective Scan (SS2D), a four-way scanning mechanism designed for spatial domain traversal.
In contrast to the self-attention mechanism (Figure~\ref{fig:att_comp} (a)), SS2D ensures that each image patch acquires contextual knowledge exclusively through a compressed hidden state computed along its corresponding scanning path (Figure~\ref{fig:att_comp} (b)), thereby reducing the computational complexity from quadratic to linear.

Building on the VSS blocks, we develop a family of VMamba architectures (\textit{i.e.}, VMamba-Tiny/Small/Base) and enhance their performance through architectural improvements and implementation optimizations.
Compared to benchmark vision models built on CNNs (ConvNeXt~\cite{liu2022convnet}), ViTs (Swin~\cite{Swin2021}, HiViT~\cite{hivit}), and SSMs (S4ND~\cite{s4ndnguyen2022s4nd}, Vim~\cite{zhu2024vision}), VMamba consistently achieves higher image classification accuracy on ImageNet-1K~\cite{ImageNet2009} across various model scales.
Specifically, VMamba-Base achieves a top-1 accuracy of $83.9\%$, surpassing Swin by $+0.4\%$, with a throughput exceeding Swin's by a substantial margin over $40\%$ ($646$ \textit{vs.} $458$).
VMamba's superiority extends across multiple downstream tasks, with VMamba-Tiny/Small/Base achieving $47.3\%$/$48.7\%$/$49.2\%$ mAP in object detection on COCO~\cite{COCO2014} ($1\times$ training schedule). This outperforms Swin by $4.6\%$/$3.9\%$/$2.3\%$ and ConvNeXt by $3.1\%$/$3.3\%$/$2.2\%$, respectively.
As for single-scale semantic segmentation on ADE20K~\cite{zhou2017scene}, VMamba-Tiny/Small/Base achieves $47.9\%$/$50.6\%$/$51.0\%$ mIoU, which surpasses Swin by $3.4\%$/$3.0\%$/$2.9\%$ and ConvNeXt by $1.9\%$/$1.9\%$/$1.9\%$, respectively.
Furthermore, unlike ViT-based models, which experience quadratic growth in computational complexity with the number of input tokens, VMamba exhibits linear growth in FLOPs while maintaining comparable performance. This demonstrates its state-of-the-art input scalability.

The contributions of this study are summarized as follows:
\begin{itemize}
    \item We propose VMamba, an SSM-based vision backbone for visual representation learning with linear time complexity. A series of architectural and implementation improvements are adopted to enhance the inference speed of VMamba.

    \item We introduce 2D Selective Scan (SS2D) to bridge 1D array scanning and 2D plane traversal, enabling the extension of selective SSMs to process vision data.
    
    \item VMamba achieves promising performance across various visual tasks, including image classification, object detection, and semantic segmentation. It also exhibits remarkable adaptability w.r.t. the length of the input sequence, showcasing linear growth in computational complexity.
\end{itemize}

\section{Related Work}\label{sec:related_work}

\paragraph{Convolutional Neural Networks (CNNs).}
Since AlexNet~\cite{AlexNet2012}, considerable efforts have been devoted to enhancing the modeling capabilities~\cite{vgg, googlenet, Resnet2016, densenet} and computational efficiency~\cite{howard2017mobilenets, EfficientNet2019, yang2021focal, radosavovic2020designing} of CNN-based models across various visual tasks. 
Sophisticated operators like depth-wise convolution~\cite{howard2017mobilenets} and deformable convolution~\cite{dai2017deformable,zhu2019deformable} have been introduced to increase the flexibility and efficacy of CNNs.
Recently, inspired by the success of Transformers~\cite{vaswani2017attention}, modern CNNs~\cite{liu2022convnet} have shown promising performance by integrating long-range dependencies~\cite{ding2022scaling,rao2022hornet,liu2023more} and dynamic weights~\cite{han2021connection} into their designs.

\paragraph{Vision Transformers (ViTs).} 
As a pioneering work, ViT~\cite{ViT2021} explores the effectiveness of vision models based on vanilla Transformer architecture, highlighting the importance of large-scale pre-training for image classification performance.
To reduce ViT's dependence on large datasets, DeiT~\cite{DeiT2021} introduces a teacher-student distillation strategy, transferring knowledge from CNNs to ViTs and emphasizing the importance of inductive bias in visual perception.
Following this approach, subsequent studies propose hierarchical ViTs~\citep{Swin2021, dong2022cswin, pvt, container, hivit, tian2023integrally, dai2021coatnet, ding2022davit, zhao2022graformer, xcit}.

Another research direction focuses on improving the computational efficiency of self-attention, which serves as the cornerstone of ViTs.
Linear Attention~\cite{katharopoulos2020transformers} reformulates self-attention as a linear dot-product of kernel feature maps, using the associativity property of matrix products to reduce computational complexity from quadratic to linear.
GLA~\cite{yang2023gated} introduces a hardware-efficient variant of linear attention that balances memory movement with parallelizability.
RWKV~\cite{rmkvpeng2023rwkv} also leverages the linear attention mechanism to combine parallelizable transformer training with the efficient inference of recurrent neural networks (RNNs).
RetNet~\cite{retnetsun2023retentive} adds a gating mechanism to enable a parallelizable computation path, offering an alternative to recurrence.
RMT~\cite{fan2024rmt} further extends this for visual representation learning by applying the temporal decay mechanism to the spatial domain.

\paragraph{State Space Models (SSMs). } 
Despite their widespread adoption in vision tasks, ViT architectures face significant challenges due to the quadratic complexity of self-attention, especially when handling long input sequences (\textit{e.g.}, high-resolution images).
In efforts to improve scaling efficiency~\cite{dao2022flashattention,dao2023flashattention, rmkvpeng2023rwkv, retnetsun2023retentive, megama2022mega}, SSMs have emerged as compelling alternatives to Transformers, attracting significant research attention.
Gu \textit{et al.}~\citep{lsslgu2021} demonstrate the potential of SSM-based models in handling the long-range dependencies using the HiPPO initialization~\citep{hippogu2020}.
To improve practical feasibility, S4~\cite{s4gu2021} proposes normalizing the parameter matrices into a diagonal structure.
Various structured SSM models have since emerged, each offering distinct architectural enhancements, such as complex-diagonal structures~\cite{dssgupta2022, s4dgu2022}, support for multiple-input multiple-output~\cite{s5smith2022simplified}, diagonal plus low-rank decomposition~\cite{liquids4hasani2022liquid}, and selection mechanisms~\cite{mambagu2023mamba}. 
These advancements have also been integrated into larger representation models~\cite{gssmehta2023long, megama2022mega, h3fu2022hungry}, further highlighting the versatility and scalability of structured state space models in various applications.
While these models primarily target long-range and sequential data such as text and speech, limited research has explored applying SSMs to vision data with two-dimensional structures.
%

\section{Preliminaries}\label{sec:preliminaries}

\paragraph{Formulation of SSMs.}\label{sec:formulation_of_ssms}
Originating from the Kalman filter~\cite{Kalman1960filter}, SSMs are linear time-invariant (LTI) systems that map the input signal $u(t) \in \mathbb{R}$ to the output response $y(t) \in \mathbb{R}$ via the hidden state $\mathbf{h}(t) \in \mathbb{R}^{N}$.
Specifically, continuous-time SSMs can be expressed as linear ordinary differential equations (ODEs) as follows,
\begin{equation}\label{eq:ssm}
\begin{split}
  \mathbf{h'}(t) &= \mathbf{A} \mathbf{h}(t) + \mathbf{B} u(t),  \\
  y(t)  &= \mathbf{C} \mathbf{h}(t) + D u(t),
\end{split}
\end{equation}
where $\mathbf{A} \in \mathbb{R}^{N\times N}$, $\mathbf{B}\in \mathbb{R}^{N\times 1}$, $\mathbf{C}\in \mathbb{R}^{1\times N}$, and $D \in \mathbb{R}^{1}$ are the weighting parameters.

\paragraph{Discretization of SSM.}\label{sec:discretization_of_ssm}
To be integrated into deep models, continuous-time SSMs must undergo discretization in advance.
Concretely, for the time interval $[t_a, t_b]$, the analytic solution of the hidden state variable $\mathbf{h}(t)$ at $t = t_b$ can be expressed as
\begin{equation}\label{eq:ode_slt}
    \mathbf{h}(t_b) = e^{\mathbf{A}(t_b-t_a)} \mathbf{h}(t_a) + e^{\mathbf{A}(t_b - t_a)} \int_{t_a}^{t_b} \mathbf{B}(\tau) u(\tau) e^{-\mathbf{A}(\tau-t_a)} \,d\tau.
\end{equation}
By sampling with the time-scale parameter $\boldsymbol{\Delta}$ (\textit{i.e.}, $d\tau |_{t_i}^{t_{i+1}} =\Delta_i$), $h(t_b)$ can be discretized by
\begin{equation}\label{eq:ode_dis}
    \mathbf{h}_b = e^{\mathbf{A}(\Delta_a+...+\Delta_{b-1})} \left( \mathbf{h}_a + \sum^{b-1}_{i=a} \mathbf{B}_i u_i e^{-\mathbf{A}(\Delta_a+...+\Delta_i)} \Delta_i \right),
\end{equation}
where $[a,b]$ is the corresponding discrete step interval.
Notably, this formulation approximates the result obtained by the zero-order hold (ZOH) method, which is frequently utilized in the literature of SSM-based models (please refer to Appendix~\ref{sec:appendix_discretization_of_state_space_models} for detailed proof).

\paragraph{Selective Scan Mechanism.}\label{sec:selective_scan_mechanism}
To address the limitation of LTI SSMs (Eq.~\ref{eq:ssm}) in capturing the contextual information, Gu \textit{et al.}~\cite{mambagu2023mamba} propose a novel parameterization method for SSMs, which incorporates an input-dependent selection mechanism (referred to as S6). 
However, for selective SSMs, the time-varying weighting parameters pose a challenge for efficient computation of hidden states, as convolutions cannot accommodate dynamic weights, making them inapplicable.
Nevertheless, since the recurrence relation of $h_b$ in Eq.~\ref{eq:ode_dis} can be derived, the response $y_b$ can still be efficiently computed using associative scan algorithms~\cite{blelloch1990prefix,martin2018parallelizing, s5smith2022simplified}, which has linear complexity (see Appendix~\ref{sec:appendix_derivation_of_the_recurrence_relation_of_selective_ssms} for a detailed explanation).

\section{VMamba: Visual State Space Model}\label{sec:vmamba}

\subsection{Network Architecture}\label{sec:network_architecture}

We develop VMamba in three scales: Tiny, Small, and Base (referred to as VMamba-T, VMamba-S, and VMamba-B, respectively).
An overview of the architecture of VMamba-T is illustrated in Figure~\ref{fig:vmamba} (a), and detailed configurations are provided in Appendix ~\ref{sec:appendix_details_for_vmamba_models}.
The input image $\mathbf{I}\in \mathbb{R}^{H\times W\times 3}$ is first partitioned into patches by a stem module, resulting in a 2D feature map with spatial dimension of $H/4\times W/4$.
%
Without incorporating additional positional embeddings, multiple network stages are employed to create hierarchical representations with resolutions of $H/8\times W/8$, $H/16\times W/16$, and $H/32\times W/32$.
Specifically, each stage comprises a down-sampling layer (except for the first stage), followed by a stack of Visual State Space (VSS) blocks.

The VSS blocks serve as the visual counterparts to Mamba blocks~\cite{mambagu2023mamba} (Figure~\ref{fig:vmamba} (b)) for representation learning.
The initial architecture of VSS blocks (referred to as the `vanilla VSS Block' in Figure~\ref{fig:vmamba} (c)) is formulated by replacing the S6 module.
S6 is the core of Mamba and achieves global receptive fields, dynamic weights (\textit{i.e.}, selectivity), and linear complexity. We substitute it with the newly proposed 2D-Selective-Scan (SS2D) module, and more details will be introduced in the following subsection.
To further enhance computational efficiency, we remove the entire multiplicative branch (highlighted by the red box in Figure~\ref{fig:vmamba} (c)), as the effect of the gating mechanism has already been achieved by the selectivity of SS2D. 
As a result, the improved VSS block (shown in Figure~\ref{fig:vmamba} (d)) consists of a single network branch with two residual modules, mimicking the architecture of a vanilla Transformer block~\cite{vaswani2017attention}.
All results in this paper are obtained using VMamba models built with VSS blocks in this architecture.

\begin{figure*}
\centering
\vskip 0.2in
\begin{center}
\includegraphics[width=\textwidth]{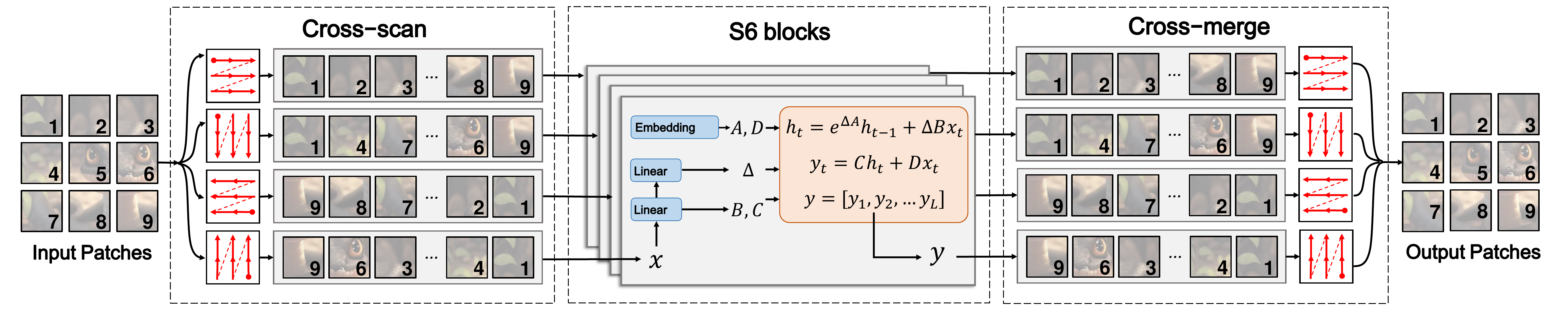}
\caption{Illustration of 2D-Selective-Scan (SS2D). Input patches are traversed along four different scanning paths (\textit{Cross-Scan}), with each sequence independently processed by separate S6 blocks. The results are then merged to construct a 2D feature map as the final output (\textit{Cross-Merge}).
}
\label{fig:ss2d}
\end{center}
\vskip -0.2in
\end{figure*}

\begin{figure*}[t]
\centering
\vskip 0.2in
\begin{center}
\includegraphics[width=\textwidth]{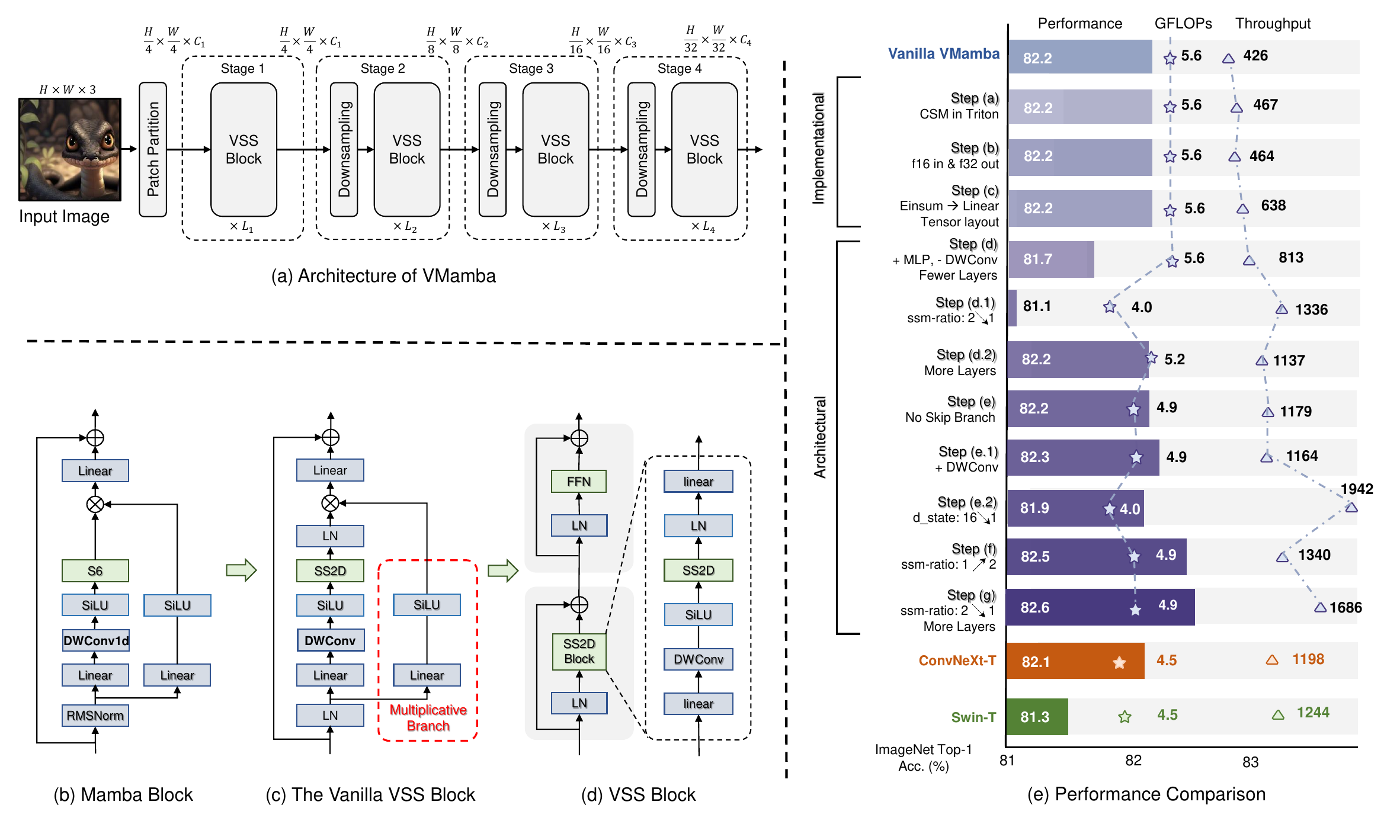}
\caption{\textbf{Left:} Illustration of (a) the overall architecture of VMamba, and (b) - (d) the structure of Mamba and VSS blocks. \textbf{Right:} Comparison of VMamba variants and benchmark methods in terms of classification accuracy and computational efficiency.
}
\label{fig:vmamba}
\end{center}
\vskip -0.2in
\end{figure*}

\subsection{2D-Selective-Scan for Vision Data (SS2D)}\label{sec:ss2d}

While the sequential nature of the scanning operation in S6 aligns well with NLP tasks involving temporal data, it poses a significant challenge when applied to vision data, which is inherently non-sequential and encompasses spatial information (\textit{e.g.}, local texture and global structure).
To address this issue, S4ND~\cite{s4ndnguyen2022s4nd} reformulates SSM with convolutional operations, directly extending the kernel from 1D to 2D through the outer-product.
However, such modification restricts the weights from being input-dependent, resulting in a limited capacity for capturing contextual information.
Therefore, we adhere to the selective scan approach~\cite{mambagu2023mamba} for input processing and propose the 2D-Selective-Scan (SS2D) module to adapt S6 to vision data without compromising its advantages.

Figure~\ref{fig:ss2d} illustrates that data forwarding in SS2D consists of three steps: cross-scan, selective scanning with S6 blocks, and cross-merge.
Specifically, SS2D first unfolds the input patches into sequences along four distinct traversal paths (\textit{i.e.}, Cross-Scan). Each patch sequence is then processed in parallel using a separate S6 block, and the resultant sequences are reshaped and merged to form the output map (\textit{i.e.}, Cross-Merge).
Through the use of complementary 1D traversal paths, SS2D allows each pixel in the image to integrate information from all other pixels across different directions. This integration facilitates the establishment of global receptive fields in the 2D space.

\subsection{Accelerating VMamba}\label{sec:accelerating_vmamba}

As shown in Figure~\ref{fig:vmamba} (e), the VMamba-T model with vanilla VSS blocks (referred to as `Vanilla VMamba') achieves a throughput of $426$ images/s and contains $22.9$M parameters with $5.6$G FLOPs.
Despite achieving a state-of-the-art classification accuracy of 82.2\% (outperforming Swin-T~\cite{Swin2021} by 0.9\% at the tiny level), the low throughput and high memory overhead present significant challenges for the practical deployment of VMamba.

In this subsection, we outline our efforts to enhance its inference speed, primarily focusing on improvements in both implementation details and architectural design. 
We evaluate the models with image classification on ImageNet-1K.
The impact of each progressive improvement is summarized as follows, 
where ($\%$, img/s) denote the gains in top-1 accuracy on ImageNet-1K and inference throughput, respectively.
Further discussion is provided in Appendix~\ref{sec:appendix_details_for_vmamba_models}.

\begin{enumerate}
    \item[Step (a)] ($+0.0\%$, $+41$~img/s) by re-implementing Cross-Scan and Cross-Merge in \texttt{Triton}.
    
    \item[Step (b)] ($+0.0\%$, $-3$~img/s) by adjusting the CUDA implementation of selective scan to accommodate \texttt{float16} input and \texttt{float32} output. This remarkably enhances the training efficiency (throughput from  $165$ to $184$), despite slight speed fluctuation at test time.
    
    \item[Step (c)] ($+0.0\%$, $+174$~img/s) by substituting the relatively slow \texttt{einsum} in selective scan with a linear transformation (\textit{i.e.}, \texttt{torch.nn.functional.linear}). 
    We also adopt the tensor layout of \texttt{(B, C, H, W)} to eliminate unnecessary data permutations.

    \item[Step (d)] ($-0.6\%$, $+175$~img/s) by introducing \texttt{MLP} into VMamba due to its computational efficiency. We also discard the \texttt{DWConv} (depth-wise convolutional~\cite{han2021connection}) layers and change the layer configuration from \texttt{[2,2,9,2]} to \texttt{[2,2,2,2]} to lower FLOPs.

    \item[Step (e)] ($+0.6\%$, $+366$~img/s) by reducing the parameter \texttt{ssm-ratio} (the feature expansion factor) from $2.0$ to $1.0$ (also referred to as Step (d.1)), raising the layer numbers to \texttt{[2,2,5,2]} (also referred to as Step (d.2)), and discarding the entire multiplicative branch as illustrated in Figure~\ref{fig:vmamba} (c).
    
    \item[Step (f)] ($+0.3\%$, $+161$~img/s) by introducing the \texttt{DWConv} layers (also referred to as Step (e.1)) and reducing the parameter \texttt{d\_state} (the SSM state dimension) from $16.0$ to $1.0$ (also referred to as Step (e.2)), together with raising \texttt{ssm-ratio} back to $2.0$. 

    \item[Step (g)] ($+0.1\%$, $+346$~img/s) by reducing the \texttt{ssm-ratio} to $1.0$ while changing the layer configuration from \texttt{[2,2,5,2]} to \texttt{[2,2,8,2]}.
    
\end{enumerate}

\section{Experiments}\label{sec:experiments}

In this section, we present a series of experiments to evaluate the performance of VMamba and compare it to popular benchmark models across various visual tasks. 
We also validate the effectiveness of the proposed 2D feature map traversal method by comparing it with alternative approaches.
Additionally, we analyze the characteristics of VMamba by visualizing its effective receptive field (ERF) and activation map, and examining its scalability with longer input sequences. 
We primarily follow the hyperparameter settings and experimental configurations used in Swin~\cite{Swin2021}. 
For detailed experiment settings, please refer to Appendix~\ref{sec:appendix_details_for_vmamba_models} and~\ref{sec:details_for_vmamba_on_downstream_tasks}, and for additional ablations, see Appendix~\ref{sec:appendix_ablations}.
All experiments were conducted on a server with 8 $\times$ NVIDIA Tesla-A100 GPUs.

\subsection{Image Classification}\label{sec:image_classification}

We evaluate VMamba's performance in image classification on ImageNet-1K~\cite{ImageNet2009}, with comparison results against benchmark methods summarized in Table~\ref{exp:imagenet-system}.
With similar FLOPs, VMamba-T achieves a top-1 accuracy of $82.6\%$, outperforming DeiT-S by $2.8\%$ and Swin-T by $1.3\%$. 
Notably, VMamba maintains its performance advantage at both Small and Base scales.
For example, VMamba-B achieves a top-1 accuracy of $83.9\%$, surpassing DeiT-B by $2.1\%$ and Swin-B by $0.4\%$. 

In terms of computational efficiency,  
VMamba-T achieves a throughput of 1,686 images/s, which is either superior or comparable to state-of-the-art methods.
This advantage continues with VMamba-S and VMamba-B, achieving throughputs of $877$ images/s and $646$ images/s, respectively.
Compared to SSM-based models, the throughput of VMamba-T is $1.47\times$ higher than S4ND-Conv-T~\cite{s4ndnguyen2022s4nd} and $1.08\times$ higher than Vim-S~\cite{zhu2024vision}, while maintaining a clear performance lead of $0.4\%$ and $2.1\%$ over these models, respectively.

\begin{table}[t]
\small
\centering
\setlength{\tabcolsep}{1pt}
\renewcommand\arraystretch{1.0}
\caption{Performance comparison on ImageNet-1K.
Throughput values are measured with an A100 GPU and an AMD EPYC 7542 CPU, using the toolkit released by~\cite{rw2019timm}, following the protocol proposed in~\cite{Swin2021}. All images are of size $224\times 224$.
}
\label{exp:imagenet-system}
\vspace{-4pt}
\resizebox{0.49\linewidth}{!}{
    \ra{1.1}
    \setlength{\tabcolsep}{0.1cm}
    \begin{tabular}{l|ccc|c}
    \Xhline{1.0pt}
    \multirow{2}{*}{Model} & Params & FLOPs & TP. & Top-1 \\
    & (M) & (G) & (img/s) & (\%) \\
    \hline
    \multicolumn{5}{c}{\textbf{Transformer-Based}}\\
    \hline
    DeiT-S~\cite{DeiT2021} & 22M & 4.6G  & 1761&  79.8 \\
    DeiT-B~\cite{DeiT2021} & 86M & 17.5G  & 503&  81.8 \\
    \hline
    HiViT-T~\citep{hivit} & 19M & 4.6G & 1393  & 82.1 \\
    HiViT-S~\citep{hivit} & 38M & 9.1G & 712 & 83.5 \\
    HiViT-B~\citep{hivit} & 66M & 15.9G & 456  & 83.8 \\
    \hline
    Swin-T~\citep{Swin2021} & 28M & 4.5G & 1244 & 81.3 \\
    Swin-S~\citep{Swin2021} & 50M & 8.7G & 718 & 83.0 \\
    Swin-B~\citep{Swin2021} & 88M & 15.4G & 458 & 83.5 \\
    \hline
    XCiT-S24~\citep{xcit} & 48M & 9.2G & 671 & 82.6 \\ 
    XCiT-M24~\cite{xcit} & 84M & 16.2G & 423 & 82.7 \\ 
    \Xhline{1.0pt}
    \end{tabular}
    }
\resizebox{0.48\linewidth}{!}{
    \ra{1.11}
    \begin{tabular}{l|ccc|c}
    \Xhline{1.0pt}
    \multirow{2}{*}{Model} & Params & FLOPs & TP. & Top-1 \\
    & (M) & (G) & (img/s) & (\%) \\
    \hline
    \multicolumn{5}{c}{\textbf{ConvNet-Based}}\\
    \hline
    ConvNeXt-T~\citep{liu2022convnet} & 29M & 4.5G & 1198  & 82.1 \\
    ConvNeXt-S~\citep{liu2022convnet} & 50M & 8.7G & 684  & 83.1 \\
    ConvNeXt-B~\citep{liu2022convnet} & 89M & 15.4G & 436  & 83.8 \\
    \hline
    \multicolumn{5}{c}{\textbf{SSM-Based}}\\
    \hline
    S4ND-Conv-T~\cite{s4ndnguyen2022s4nd} & 30M & 5.2G  & 683 & 82.2 \\
    S4ND-ViT-B~\cite{s4ndnguyen2022s4nd} & 89M & 17.1G  & 397 & 80.4 \\
    Vim-S~\cite{zhu2024vision} & 26M & 5.3G  & 811 & 80.5 \\
    \hline
    \rowcolor{gray!20}
    VMamba-T & 30M & 4.9G  & 1686 & 82.6 \\
    \rowcolor{gray!20}
    VMamba-S & 50M & 8.7G & 877 & 83.6 \\
    \rowcolor{gray!20}
    VMamba-B & 89M & 15.4G & 646 & 83.9 \\
    \Xhline{1.0pt}
    \end{tabular}
    }
    \vspace{-8pt}
\end{table}

\subsection{Downstream Tasks}\label{sec:downstream_tasks}
In this sub-section, we evaluate the performance of VMamba on downstream tasks, including object detection and instance segmentation on MSCOCO2017~\cite{COCO2014}, and semantic segmentation on ADE20K~\cite{zhou2017scene}.
The training framework is based on the MMDetection~\cite{MMdet2019} and MMSegmenation~\cite{mmseg2020} libraries, following~\cite{liu2022swin} in utilizing Mask R-CNN~\cite{MaskRCNN2017} and UperNet~\cite{upernet} as the detection and segmentation networks, respectively.

\begin{table}[t]
\small
\centering
\setlength{\tabcolsep}{1pt}
\renewcommand\arraystretch{1.0}
\caption{\textbf{Left:} Results for object detection and instance segmentation on MSCOCO. $AP^{b}$ and $AP^{m}$ denote box AP and mask AP, respectively. FLOPs are calculated with an input size of $1280\times800$.
The notation `$1\times$' indicates models fine-tuned for $12$ epochs, while `$3\times$MS' denotes multi-scale training for $36$ epochs.
\textbf{Right:} Results for semantic segmentation on ADE20K. FLOPs are calculated with an input size of $512\times2048$. `SS' and `MS' denote single-scale and multi-scale testing, respectively.
}
\label{exp:coco_ade20k}

\resizebox{0.46\textwidth}{!}{
    \ra{1.19}
    \setlength{\tabcolsep}{0.2cm}
    \begin{tabular}{l|cc|cc}
    \Xhline{1.0pt}
    \multicolumn{5}{c}{\textbf{Mask R-CNN 1$\times$ schedule}}\\
    \hline
    Backbone & AP$^\text{b}$ & AP$^\text{m}$ & Params & FLOPs \\
    \hline
    Swin-T & 42.7  & 39.3  & 48M & 267G \\
    ConvNeXt-T & 44.2  & 40.1  & 48M & 262G \\
    \rowcolor{gray!20}
    VMamba-T & 47.3  & 42.7 & 50M & 271G \\
    \hline
    Swin-S & 44.8 & 40.9   & 69M & 354G \\
    ConvNeXt-S & 45.4  & 41.8  & 70M & 348G \\
    \rowcolor{gray!20}
    VMamba-S & 48.7  & 43.7  & 70M & 349G \\
    \hline
    Swin-B & 46.9  & 42.3  & 107M & 496G \\
    ConvNeXt-B & 47.0  & 42.7  & 108M & 486G \\
    \rowcolor{gray!20}
    VMamba-B & 49.2  & 44.1  & 108M & 485G \\
    \hline
    \multicolumn{5}{c}{\textbf{Mask R-CNN 3$\times$ MS schedule}}\\
    \hline
    Swin-T &  46.0  & 41.6  & 48M & 267G \\
    ConvNeXt-T &  46.2  & 41.7  & 48M & 262G \\
    NAT-T & 47.7 & 42.6 & 48M & 258G \\
    \rowcolor{gray!20}
    VMamba-T & 48.8  & 43.7  & 50M & 271G \\
    \hline
    Swin-S & 48.2  & 43.2  & 69M & 354G \\
    ConvNeXt-S & 47.9  & 42.9 & 70M & 348G \\
    NAT-S & 48.4 & 43.2 & 70M & 330G \\
    \rowcolor{gray!20}
    VMamba-S & 49.9  & 44.2 & 70M & 349G \\
    \Xhline{1.0pt}
    \end{tabular}
}
\resizebox{0.52\textwidth}{!}{
    \ra{1.12}
    \setlength{\tabcolsep}{0.2cm}
    \begin{tabular}{l|cc|cc}
    \Xhline{1.0pt}
    \multicolumn{5}{c}{\textbf{ADE20K with crop size 512}}\\
    \hline
    Backbone & \begin{tabular}[c]{@{}c@{}} mIOU \\ (SS) \end{tabular} &  \begin{tabular}[c]{@{}c@{}}mIOU \\ (MS) \end{tabular}  & Params & FLOPs \\
    \hline
    ResNet-50 & 42.1 & 42.8 & 67M & 953G  \\
    DeiT-S + MLN & 43.8 & 45.1 & 58M & 1217G \\
    Swin-T & 44.5 & 45.8 & 60M & 945G \\
    ConvNeXt-T & 46.0 & 46.7 & 60M & 939G \\
    NAT-T & 47.1 & 48.4 & 58M & 934G \\
    Vim-S & 44.9 & - & 46M & - \\
    \rowcolor{gray!20}
    VMamba-T & 47.9 & 48.8 & 62M & 949G  \\
    \hline
    ResNet-101 & 43.8 & 44.9 & 86M & 1030G  \\
    DeiT-B + MLN & 45.5 & 47.2 & 144M & 2007G \\
    Swin-S & 47.6 & 49.5 & 81M & 1039G \\ 
    ConvNeXt-S & 48.7 & 49.6 & 82M & 1027G \\
    NAT-S & 48.0 & 49.5 & 82M & 1010G \\
    \rowcolor{gray!20}
    VMamba-S & 50.6 & 51.2 & 82M & 1028G  \\
    \hline
    Swin-B & 48.1 & 49.7 & 121M & 1188G \\
    ConvNeXt-B & 49.1 & 49.9 & 122M & 1170G \\
    NAT-B & 48.5 & 49.7 & 123M & 1137G \\
    RepLKNet-31B &  49.9 & 50.6 & 112M & 1170G \\
    \rowcolor{gray!20}
    VMamba-B & 51.0 & 51.6 & 122M & 1170G  \\
    \Xhline{1.0pt}
    \end{tabular}
}
\vspace{-8pt}
\end{table}

\paragraph{Object Detection and Instance Segmentation.}\label{sec:object_detection}
The results on MSCOCO are presented in Table~\ref{exp:coco_ade20k}. 
VMamba demonstrates superior performance in both box and mask Average Precision (AP$^\text{b}$ and AP$^\text{m}$) across different training schedules.
Under the $12$-epoch fine-tuning schedule, VMamba-T/S/B achieves object detection mAPs of $ 47.3\%/48.7\%/49.2\%$, outperforming Swin-T/S/B by $4.6\%/3.9\%/2.3\%$ mAP and ConvNeXt-T/S/B by $3.1\%/3.3\%/2.2\%$ mAP, respectively.
VMamba-T/S/B achieves instance segmentation mAPs that exceed Swin-T/S/B by $3.4\%/2.8\%/$1.8$\%$ mAP and ConvNeXt-T/S/B by $2.6\%/1.9\%/$1.4$\%$ mAP, respectively.
Furthermore, VMamba's advantages persist with the 36-epoch fine-tuning schedule using multi-scale training, highlighting its strong potential in downstream tasks requiring dense predictions.

\paragraph{Semantic Segmentation.}\label{sec:semantic_segmentation}
Consistent with previous experiments, VMamba demonstrates superior performance in semantic segmentation on ADE20K with a comparable amount of parameters.
As shown in Table~\ref{exp:coco_ade20k}, VMamba-T achieves $3.4\%$ higher mIoU than Swin-T and $1.9\%$ higher than ConvNeXt-T in the Single-Scale (SS) setting, and the advantage persists with Multi-Scale (MS) input.
For models at the Small and Base levels, VMamba-S/B outperforms NAT-S/B~\cite{hassani2023neighborhood} by $2.6\%$/$2.5\%$ mIoU in the SS setting, and $1.7\%$/$1.9\%$ mIoU in the MS setting. 

\begin{figure*}
\centering
\vskip 0.2in
\begin{center}
\includegraphics[width=\textwidth]{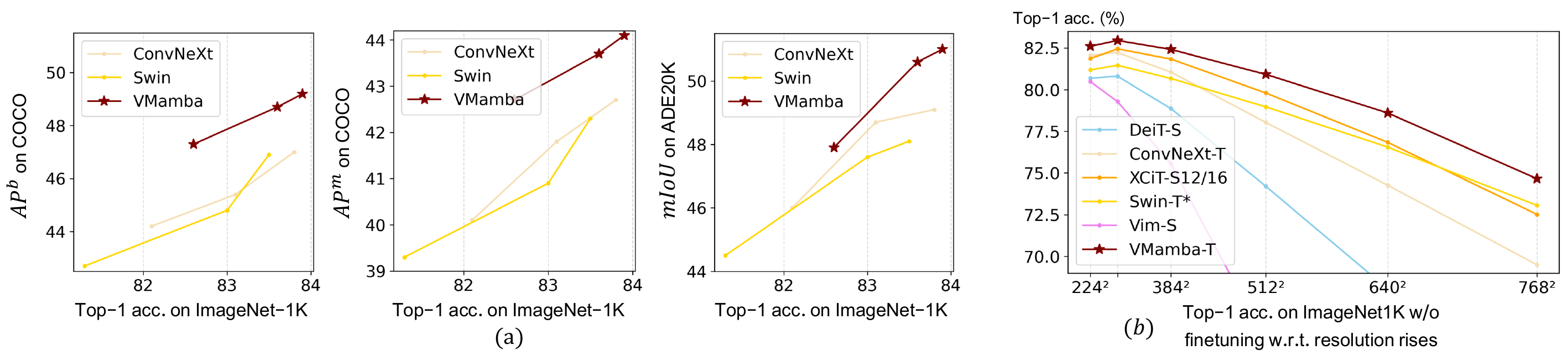}
\caption{Illustration of VMamba's adaptability to (a) downstream tasks and (b) input images with progressively increasing resolutions. Swin-T$^*$ denotes Swin-T tested with scaled window sizes.
} 
\label{fig:scaleup}
\end{center}
\vskip -0.2in
\end{figure*}

\begin{figure*}
\centering
\vskip 0.2in
\begin{center}
\includegraphics[width=\textwidth]{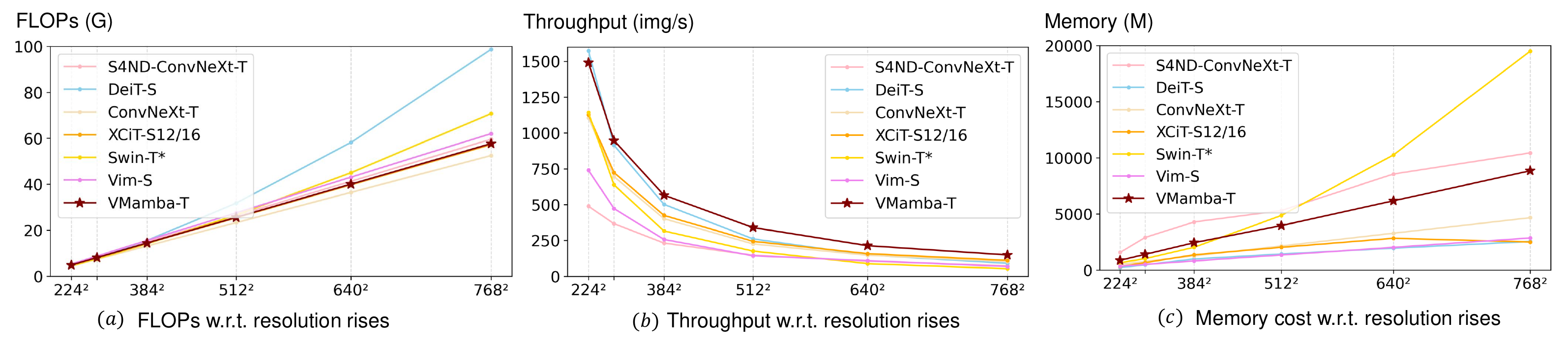}
\caption{Illustration of VMamba's resource consumption with progressively increasing resolutions. Swin-T$^*$ denotes Swin-T tested with scaled window sizes.
} 
\label{fig:scaleupcost}
\end{center}
\vskip -0.2in
\end{figure*}

\paragraph{Discussion}\label{sec:downstream_discussion}
The experimental results in this subsection demonstrate VMamba's adaptability to object detection, instance segmentation, and semantic segmentation. 
In Figure~\ref{fig:scaleup} (a), we compare VMamba's performance with Swin and ConvNeXt, highlighting its advantages in handling downstream tasks with comparable classification accuracy on ImageNet-1K.
This result aligns with Figure~\ref{fig:scaleup} (b), where VMamba shows the most stable performance (\textit{i.e.}, modest performance drop) across different input image sizes, achieving a top-1 classification accuracy of $74.7\%$ without fine-tuning ($79.2\%$ with \texttt{linear tuning}) at an input resolution of $768\times768$.
While exhibiting greater tolerance to changes in input resolution, VMamba also maintains linear growth in FLOPs and memory-consumption (see Figure~\ref{fig:scaleupcost} (a) and (c)) and maintains high throughput ( Figure~\ref{fig:scaleupcost} (b)),
making it more effective and efficient compared to ViT-based methods when adapting to downstream tasks with inputs of larger spatial resolutions.
This aligns with Mamba's advanced capability in efficient long sequence modeling~\cite{mambagu2023mamba}.

\subsection{Analysis}\label{sec:analysis_and_discussion}

\begin{figure*}
\centering
\vskip 0.2in
\begin{center}
\includegraphics[width=\textwidth]{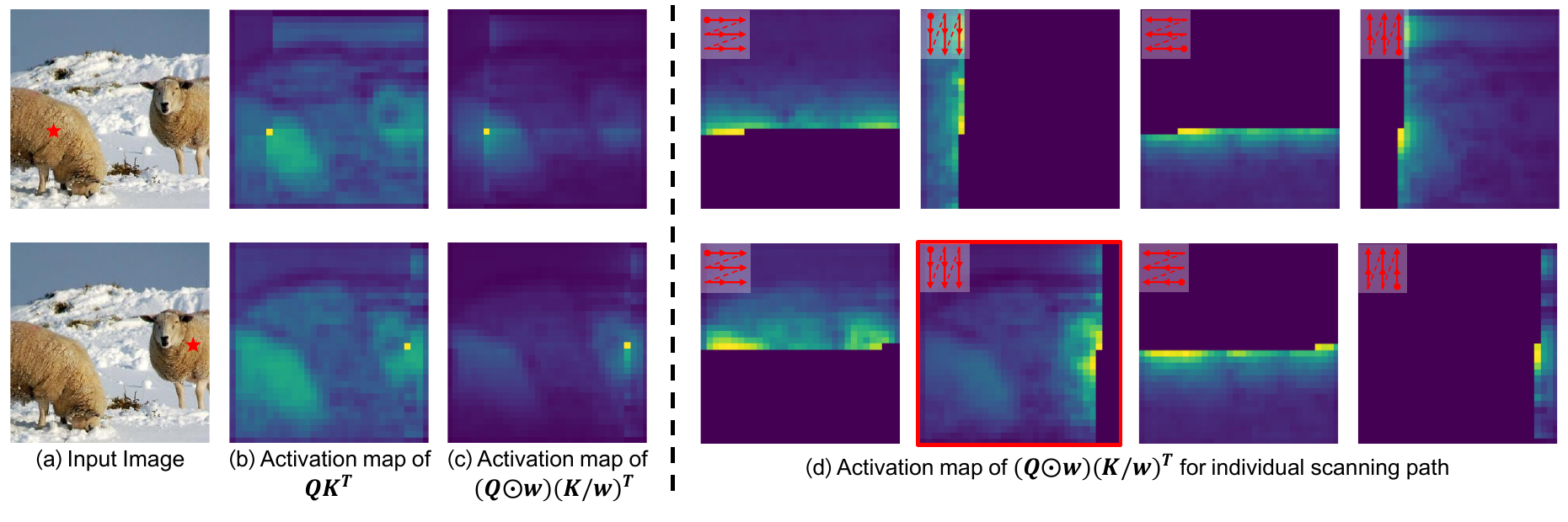}
\caption{Illustration of the activation map for query patches indicated by red stars. The visualization results in (b) and (c) are obtained by combining the activation maps from each scanning path in SS2D.
}
\label{fig:attention}
\end{center}
\vskip -0.2in
\end{figure*}

\paragraph{Relationship between SS2D and Self-Attention.}\label{sec:relationship_between_ss2d and_self_attention} 
To formulate the response $\mathbf{Y}$ within the time interval $[a, b]$ of length $T$, 
we denote the corresponding SSM-related variables 
$\mathbf{u}_i \odot \boldsymbol{\Delta}_i\in \mathbb{R}^{1\times D_v}$, $\mathbf{B}_i\in \mathbb{R}^{1\times D_k}$, and $\mathbf{C}_i\in \mathbb{R}^{1\times D_k}$ 
as 
$\mathbf{V} \in \mathbb{R}^{T\times D_v}$, $\mathbf{K} \in \mathbb{R}^{T\times D_k}$, and $\mathbf{Q} \in \mathbb{R}^{T\times D_k}$, respectively.
Therefore, the $j\text{-th}$ slice along dimension $D_v$ of $\mathbf{y_b}$, denoted as $\mathbf{y_b}^{(j)}\in\mathbb{R}$ can be written as
\begin{align}\label{eq:ode_slt_yj}
    \mathbf{y_b}^{(j)} = \left( \mathbf{Q_T} \odot \mathbf{w_T}^{(j)} \right) 
{\mathbf{h_a}^{(j)}} +  \mathbf{Q_T} \sum_{i=1}^{T} \left(\frac{\mathbf{w_T}^{(j)}}{\mathbf{w_i}^{(j)}}  \odot \mathbf{K_i}\right)^\top \odot \left(\mathbf{V_i}^{(j)}\right).
\end{align}
where $\mathbf{h_a} \in \mathbb{R}^{D_k}$ is the hidden state at step $a$, $\odot$ denotes element-wise product. Particularly, $\mathbf{V_i}^{(j)}$ is only a scalar. The formulation of each element in $\mathbf{w}\coloneqq \left[\mathbf{w}_1; \ldots; \mathbf{w}_T\right]\in\mathbb{R}^{T\times D_k\times D_v}$, \textit{i.e.}, $\mathbf{w}_i\in \mathbb{R}^{D_k\times D_v}$, can be written as $\mathbf{w}_i=\prod_{j=1}^{i} e^{\mathbf{A}\boldsymbol{\Delta}_{a-1+j}^\top}$, representing the cumulative attention weight at step $i$ computed along the scanning path.

Consequently, the $j\text{-th}$ dimension of $\mathbf{Y}$, \textit{i.e.}, $\mathbf{Y}^{(j)}\in\mathbb{R}^{T\times 1}$, can be expressed as 
\begin{equation}\label{eq:VMamba_att}
    \mathbf{Y}^{(j)} = \left[\mathbf{Q} \odot \mathbf{w}^{(j)}\right] \mathbf{h_a}^{(j)} + \left[ \left( \mathbf{Q} \odot \mathbf{w}^{(j)} \right) \left( \frac{\mathbf{K}}{\mathbf{w}^{(j)}} \right)^\top \odot \mathbf{M} \right] \mathbf{V}^{(j)},
\end{equation}
where $\mathbf{M}$ denotes the temporal mask matrix of size $T\times T$ with the lower triangular part set to 1 and elsewhere 0.
Please refer to Appendix~\ref{sec:details_for_the_relationship_between_ss2d_and_self_attention} for more detailed derivations.

In Eq.~\ref{eq:VMamba_att}, the matrix multiplication process involving $\mathbf{Q}$, $\mathbf{K}$, and $\mathbf{V}$ closely resembles the self-attention mechanism, despite the inclusion of $\mathbf{w}$.

\paragraph{Visualization of Activation Maps.}\label{sec:visualization_of_attention_maps}

To gain an intuitive and in-depth understanding of SS2D, we further visualize the attention values in $\mathbf{Q} \mathbf{K}^\top$ and $\left(\mathbf{Q} \odot \mathbf{w}\right) \left(\mathbf{K} / \mathbf{w}\right)^\top$ corresponding to a specific query patch within foreground objects (referred to as the `activation map').
As shown in Figure~\ref{fig:attention} (b), the activation map of $\mathbf{Q} \mathbf{K}^\top$ demonstrates the effectiveness of SS2D in capturing and retaining traversed information, with all previously scanned tokens in the foreground region being activated.
Furthermore, the inclusion of $\mathbf{w}$ results in activation maps that are more focused on the neighborhood of query patches (Figure~\ref{fig:attention} (c)), which is consistent with the temporal weighting effect inherent in the formulation of $\mathbf{w}$.
Nevertheless, the selective scan mechanism allows VMamba to accumulate history along the scanning path, facilitating the establishment of long-term dependencies across image patches.
This is evident in the sub-figure encircled by a red box (Figure~\ref{fig:attention} (d)), where patches of the sheep far to the left (scanned in earlier steps) remain activated.
For more visualizations and further discussion, please refer to Appendix~\ref{sec:activation_map_visualization}.

\paragraph{Visualization of Effective Receptive Fields.}\label{sec:visualization_of_effective receptive_fields}

\begin{figure*}
    \centering
    \vskip 0.2in
    \begin{center}
    \includegraphics[width=\textwidth]{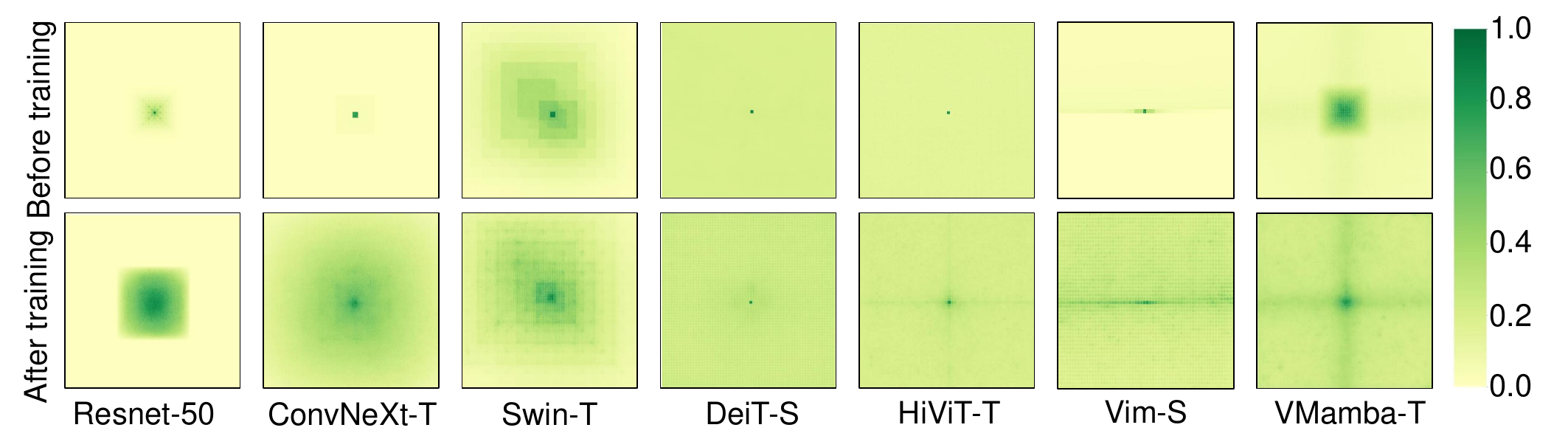}
    \caption{Comparison of Effective Receptive Fields (ERF)~\cite{luo2016understanding} between VMamba and other benchmark models. Pixels with higher intensity indicate larger responses related to the central pixel.
    }
    \label{fig:erf}
    \end{center}
    \vskip -0.2in
\end{figure*}

The Effective Receptive Field (ERF)~\cite{luo2016understanding, ding2022scaling} refers to the region in the input space that contributes to the activation of a specific output unit.
We conduct a comparative analysis of the central pixel's ERF across various visual backbones, both before and after training.
%
%
The results presented in Figure~\ref{fig:erf} illustrate that among the models examined, only DeiT, HiViT, Vim and VMamba demonstrate global ERFs, while the others exhibit local ERFs despite their theoretical potential for global coverage.
Moreover, VMamba's linear time complexity enhances its computational efficiency compared to DeiT and HiViT, which incur quadratic costs w.r.t. the number of input patches.
%
%
While both VMamba and Vim are based on the Mamba architecture, VMamba's ERF is more uniform and 2D-aware than that of Vim, which may intuitively explain its superior performance.

\begin{figure*}    
    \centering
    \vskip 0.2in
    \begin{center}
    \includegraphics[width=\textwidth]{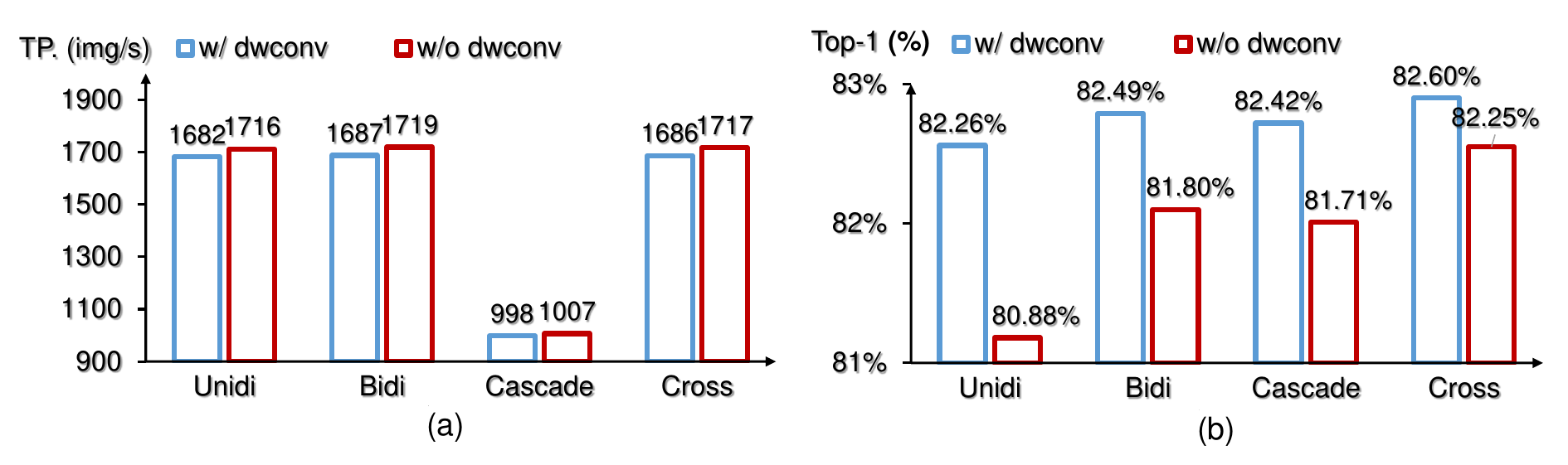}
    \caption{Performance comparison of different scanning patterns. The proposed Cross-Scan achieves superior performance in speed while maintaining the same number of parameters and FLOPs.
    }
    \label{fig:ablate_csm}
    \end{center}
    \vskip -0.2in
\end{figure*}

\paragraph{Diagnostic Study on Selective Scan Patterns.}\label{sec:diagnostic_study_on_ selective_scan_patterns}

We compare the proposed scanning pattern (\textit{i.e.} Cross-Scan) to three benchmark patterns: unidirectional scanning (Unidi-Scan), bidirectional scanning (Bidi-Scan), and cascade scanning (Cascade-Scan, scanning the data row-wise and column-wise successively).
Feature dimensions are adjusted to maintain similar architectural parameters and FLOPs for a fair comparison.
As illustrated in Figure~\ref{fig:ablate_csm}, Cross-Scan outperforms the other scanning patterns in both computational efficiency and classification accuracy, highlighting its effectiveness in achieving 2D-Selective-Scan.
Removing the \texttt{DWConv} layer, which has been shown to aid the model in learning 2D spatial information, further enhances this advantage.
This underscores the inherent strength of Cross-Scan in capturing 2D contextual information through its adoption of four-way scanning.

\section{Conclusion}\label{sec:conclusion_and_limitation}

This paper presents VMamba, an efficient vision backbone model built with State Space Models (SSMs). 
VMamba integrates the advantages of selective SSMs from NLP tasks into visual data processing, bridging the gap between ordered 1D scanning and non-sequential 2D traversal through the novel SS2D module.
Furthermore, we have significantly improved the inference speed of VMamba through a series of architectural and implementation refinements.
The effectiveness of the VMamba family has been demonstrated through extensive experiments, and its linear time complexity makes VMamba advantageous for downstream tasks with large-resolution inputs.

\paragraph{Limitations.} 

While VMamba demonstrates promising experimental results, there is still room for improvement in this study.
Previous research has validated the efficacy of unsupervised pre-training on large-scale datasets (\textit{e.g.}, ImageNet-21K). 
However, the compatibility of existing pre-training methods with SSM-based architectures like VMamba, as well as the identification of pre-training techniques specifically tailored for such models, remain unexplored.
Investigating these aspects could serve as a promising avenue for future research in architectural design.
Additionally, limited computational resources have prevented us from exploring VMamba's architecture at the Large scale and conducting a fine-grained hyperparameter search to further enhance experimental performance.
Although SS2D, the core component of VMamba, does not make specific assumptions about the layout or modality of the input data, allowing it to generalize across various tasks, the potential of VMamba for integration into more generalized tasks remains unexplored. Bridging the gap between SS2D and these tasks, along with proposing a more generalized scanning pattern for vision tasks, represents a promising research direction.

\section{Acknowledgments}
This work was supported by 
National Natural Science Foundation of China (NSFC) under Grant No.62225208 and 62406304, 
CAS Project for Young Scientists in Basic Research under Grant No.YSBR-117, 
China Postdoctoral Science Foundation under Grant No.2023M743442, 
and Postdoctoral Fellowship Program of CPSF under Grant No.GZB20240730.

\newpage
{
\small
\bibliographystyle{plain}
\bibliography{reference}
}

\newpage
\appendix
\newpage

\section{Discretization of State Space Models (SSMs)}\label{sec:appendix_discretization_of_state_space_models}

In this section, we explore the correlation between the discretized formulations of State Space Models (SSMs) obtained in Sec.~\ref{sec:preliminaries} and those derived from the zero-order hold (ZOH) method~\cite{mambagu2023mamba}, which is frequently used in studies related to SSMs.

Recall the discretized formulation of SSMs derived in Sec.~\ref{sec:preliminaries} as follows, 
\begin{equation}\label{eq:ode_dis_app}
    \mathbf{h}_b = e^{\mathbf{A}\left(\Delta_a+ \ldots +\Delta_{b-1}\right)} \left( \mathbf{h}_a + \sum^{b-1}_{i=a} \mathbf{B}_i u_i e^{-\mathbf{A}\left(\Delta_a+\ldots+\Delta_i\right)} \Delta_i \right).
\end{equation}
Let $b=a+1$, then the above equation can be re-written as
\begin{align}\label{eq:ode_dis_step_app}
      \mathbf{h}_{a+1} &= e^{\mathbf{A}\Delta_a} \mathbf{h}_a +  \mathbf{B}_a \Delta_a  u_a,
\end{align}
where $\mathbf{\overline{A_a}} \coloneq e^{\mathbf{A}\Delta_a}$ is the exact discretized form of the evolution matrix $\mathbf{A}$ obtained by ZOH, and $\mathbf{\overline{B_a}} \coloneq \mathbf{B}_a \Delta_a$ represents the first-order Taylor expansion of the discretized $\mathbf{B}$ acquired through ZOH.

\section{Derivation of the Recurrence Relation of Selective SSMs}\label{sec:appendix_derivation_of_the_recurrence_relation_of_selective_ssms}

In this section, we derive the recurrence relation of the hidden state in selective SSMs.
Given the expression of $h_b$ shown in Eq.~\ref{eq:ode_dis_app}, let us denote $e^{\mathbf{A}\left(\Delta_a+ \ldots +\Delta_{i-1}\right)}$ as $\mathbf{p_{A, a}^{i}}$. Then, its recurrence relation can be directly written as 
\begin{equation}\label{eq:recur_pA_app}
    \mathbf{p_{A, a}^{i}} = e^{\mathbf{A}\Delta_{i-1}} \mathbf{p_{A, a}^{i-1}}.
\end{equation}
For the second term of Eq.~\ref{eq:ode_dis_app}, we have
\begin{align}\label{eq:recur_pB_app}
\mathbf{p_{B, a}^{b}} &= e^{\mathbf{A}\left(\Delta_a+ \ldots + \Delta_{b-1}\right)} \sum^{b-1}_{i=a} \mathbf{B_i} u_i e^{-\mathbf{A}\left(\Delta_a+ \ldots +\Delta_i\right)} \Delta_i \\
             &= e^{\mathbf{A} \Delta_{b-1}} \mathbf{p_{B, a}^{b - 1}} + \mathbf{B_{b - 1}} u_
{b - 1} \Delta_{b-1}.
\end{align}
Therefore, with the associations derived in Eq.~\ref{eq:recur_pA_app} and Eq.~\ref{eq:recur_pB_app}, $\mathbf{h}_b=\mathbf{p_{A, a}^{b}} \mathbf{h}_a + \mathbf{p_{B, a}^{b}}$ can be efficiently computed in parallel using associative scan algorithms~\cite{blelloch1990prefix,martin2018parallelizing,s5smith2022simplified}, which are supported by numerous modern programming libraries. 
This approach effectively reduces the overall computational complexity to linear, and VMamba further accelerates the computation by adopting a hardware-aware implementation~\cite{mambagu2023mamba}.

\section{Details of the relationship between SS2D and Self-attention}\label{sec:details_for_the_relationship_between_ss2d_and_self_attention}

In this section, we clarify the relationship between SS2D and the self-attention mechanism commonly employed in existing vision backbone models. 
Subsequently, visualization results are provided to substantiate our explanation.

Let $T$ denote the length of the sequence with indices from $a$ to $b$, we define the following variables

\begin{align}
    \mathbf{V} & \coloneqq \left[\mathbf{V_{1}}; \ldots; \mathbf{V_{T}}\right]\in\mathbb{R}^{T\times D_v},  \: \text{where} \: \mathbf{V_{i}} \coloneqq  \mathbf{u_{a+i-1}} \odot \boldsymbol{\Delta_{a+i-1}}\in \mathbb{R}^{1\times D_v} \\
    \mathbf{K} & \coloneqq \left[\mathbf{K_{1}}; \ldots; \mathbf{K_{T}}\right]\in\mathbb{R}^{T\times D_k},  \: \text{where} \: \mathbf{K_{i}} \coloneqq  \mathbf{B_{a+i-1}}\in \mathbb{R}^{1\times D_k} \\
    \mathbf{Q} & \coloneqq \left[\mathbf{Q_{1}}; \ldots; \mathbf{Q_{T}}\right]\in\mathbb{R}^{T\times D_k},  \: \text{where} \: \mathbf{Q_{i}} \coloneqq  \mathbf{C_{a+i-1}}\in \mathbb{R}^{1\times D_k} \\
    \mathbf{w} & \coloneqq \left[\mathbf{w_{1}}; \ldots; \mathbf{w_{T}}\right]\in\mathbb{R}^{T\times D_k\times D_v}, \: \text{where} \: \mathbf{w_{i}} \coloneqq \prod_{j=1}^{i} e^{\mathbf{A}\boldsymbol{\Delta_{a-1+j}^\top}}\in \mathbb{R}^{D_k\times D_v} \label{eq:w}\\
    \mathbf{H} & \coloneqq \left[\mathbf{h_{a+1}}; \ldots; \mathbf{h_b}\right]\in\mathbb{R}^{T\times D_k\times D_v}, \: \text{where} \: \mathbf{h_{i}} \in \mathbb{R}^{D_k\times D_v}\\
    \mathbf{Y} & \coloneqq \left[\mathbf{y_{a+1}}; \ldots; \mathbf{y_b}\right]\in\mathbb{R}^{T\times D_v}, \: \text{where} \: \mathbf{y_{i}} \in \mathbb{R}^{D_v}
\end{align}

Note that in practice, the parameter $A$ in Eq.~\ref{eq:ssm} is simplified to $\mathbb{R}^{1 \times D_k}$. Consequently, $\mathbf{h'}(t) = \mathbf{A} \mathbf{h}(t) + \mathbf{B} u(t)$ is simplified to $\mathbf{h'}(t) = \mathbf{A} \odot \mathbf{h}(t) + \mathbf{B} u(t)$, which is the reason why  $\mathbf{w_{i}} \in\mathbb{R}^{D_k\times D_v}$.

Based on these notations, the discretized solution of time-varying SSMs (Eq.~\ref{eq:ode_dis_app}) can be written as
\begin{equation}\label{eq:ode_slt_h}
    \mathbf{h_b} = \mathbf{w_T} \odot \mathbf{h_a} + \sum_{i=1}^{T} \frac{\mathbf{w_T}}{\mathbf{w_i}} \odot \left( \mathbf{K_i}^\top \mathbf{V_i} \right),
\end{equation}
where $\odot$ denotes the element-wise product between matrices, and the division is also elements-wise.

Based on the expression of the hidden state $\mathbf{h_b}$, the first term of the output of SSM, \textit{i.e.}, $\mathbf{y_b}$, can be computed by
\begin{align}\label{eq:ode_slt_y}
    \mathbf{y_b} &= \mathbf{Q_T}\mathbf{h_b} \\
                     &= \mathbf{Q_T} \left(\mathbf{w_T} \odot 
\mathbf{h_a}\right) + \mathbf{Q_T} \sum_{i=1}^{T} \frac{\mathbf{w_T}}{\mathbf{w_i}} \odot \left( \mathbf{K_i}^\top \mathbf{V_i} \right).
\end{align}
Here, we drop the skip connection between the input and the response for simplicity.
Particularly, the $j\text{-th}$ slice along dimension $D_v$ of $\mathbf{y_b}$, denoted as $\mathbf{y_b}^{(j)}\in\mathbb{R}$ can be written as
\begin{align}\label{eq:ode_slt_yj}
    \mathbf{y_b}^{(j)} = \left( \mathbf{Q_T} \odot \mathbf{w_T}^{(j)} \right) 
{\mathbf{h_a}^{(j)}} +  \sum_{i=1}^{T} \left(\frac{\mathbf{Q_T} \odot \mathbf{w_T}^{(j)}}{\mathbf{w_i}^{(j)}} \mathbf{K_i}^\top \right) \odot \mathbf{V_i}^{(j)}.
\end{align}
Similarly, the $j\text{-th}$ slice along dimension $D_v$ of the overall response $\mathbf{Y}$, denoted as $\mathbf{Y}^{(j)}\in\mathbb{R}^{T\times 1}$, can be expressed as
\begin{equation}\label{eq:VMamba_att_detail}
    \mathbf{Y}^{(j)} = \left(\mathbf{Q} \odot \mathbf{w}^{(j)}\right) {\mathbf{h_a}^{(j)}} + \left[ \left( \mathbf{Q} \odot \mathbf{w}^{(j)} \right) \left( \frac{\mathbf{K}}{\mathbf{w}^{(j)}} \right)^\top \odot \mathbf{M} \right]\mathbf{V}^{(j)},
\end{equation}
where $\mathbf{M} \coloneqq \texttt{tril}(T, T) \in \{0,1\}^{T\times T}$ denotes the temporal mask matrix with the lower triangular portion of a $T\times T$ matrix set to 1 and elsewhere 0.
It is evident that how matrices $\mathbf{Q}$, $\mathbf{K}$, and $\mathbf{V}$ are multiplied in Eq.~\ref{eq:VMamba_att_detail} closely resembles the process in the self-attention module of Vision Transformers. 
Moreover, if $\mathbf{w}$ is in shape $(T, D_k)$ rather than $(T, D_k, D_v)$, then Eq.~\ref{eq:ode_slt_y} and Eq.~\ref{eq:VMamba_att_detail} reduce to the form of Gated Linear Attention (GLA)~\cite{yang2023gated}, indicating that GLA is also a special case of Mamba.

\section{Visualization of Attention and Activation Maps}\label{sec:activation_map_visualization}

In the preceding subsection, we illustrated how the computational process of selective SSMs shares similarities with self-attention mechanisms, allowing us to delve into the internal mechanism of SS2D through the visualization of its weight matrices.

\begin{figure*}
\centering
\vskip 0.2in
\begin{center}
\includegraphics[width=0.9\textwidth]{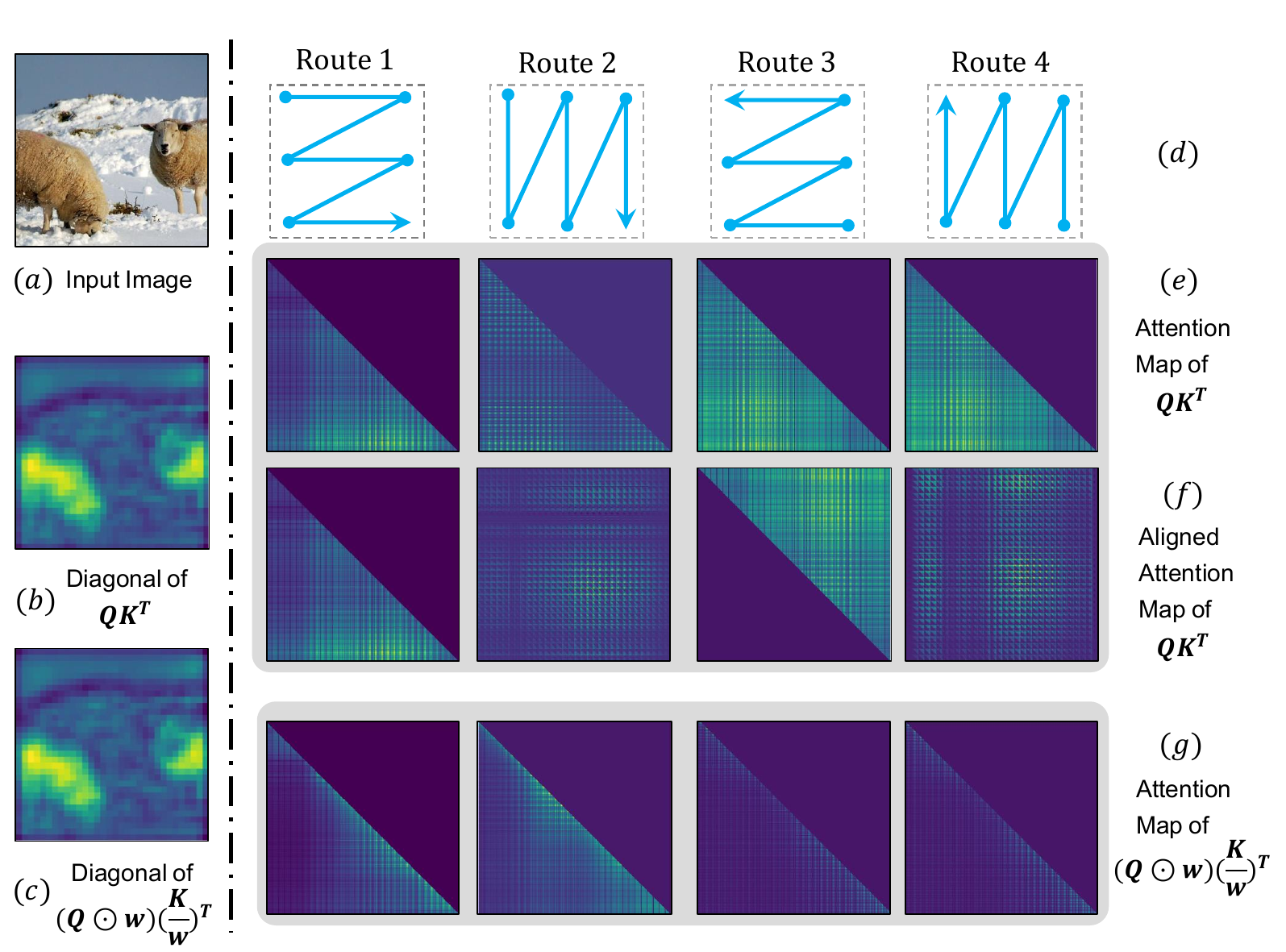}
\caption{Illustration of the attention maps obtained by SS2D.}
\label{fig:app_attention_attnmap}
\end{center}
\vskip -0.2in
\end{figure*}

Given the input image shown in Figure~\ref{fig:app_attention_attnmap} (a), illustrations of four scanning paths in SS2D are presented in Figure~\ref{fig:app_attention_attnmap} (d). The visualizations of the corresponding attention maps, calculated using $\mathbf{Q}\mathbf{K}^\top$ and $\left(\mathbf{Q} \odot \mathbf{w}\right) \left(\mathbf{K} / \mathbf{w} \right)^\top$ are shown in Figure~\ref{fig:app_attention_attnmap} (e) and Figure~\ref{fig:app_attention_attnmap} (g) respectively. 
These results underscore the effectiveness of the proposed scanning approach (\textit{i.e.}, Cross-Scan) in capturing and retaining the traversed information, as each row in a single attention map corresponds to the attention between the current patch and all previously scanned foreground tokens impartially.
Additionally, in Figure~\ref{fig:app_attention_attnmap} (f), we showcase the transformed activation maps, where the pixel order corresponds to that of the first route, traversing the image row-wise from the upper-left to the bottom-right.

By rearranging the diagonal elements of the obtained attention map in the image space, we derive the visualization results shown in Figure~\ref{fig:app_attention_attnmap} (b) and Figure~\ref{fig:app_attention_attnmap} (c) corresponding to $\mathbf{Q}\mathbf{K}^\top$ and $\left(\mathbf{Q} \odot \mathbf{w}\right) \left(\mathbf{K} / \mathbf{w} \right)^\top$ respectively. 
These maps illustrate the effectiveness of VMamba in accurately distinguishing between foreground and background pixels within an image.

\begin{figure*}
\centering
\vskip 0.2in
\begin{center}
\includegraphics[width=0.99\textwidth]{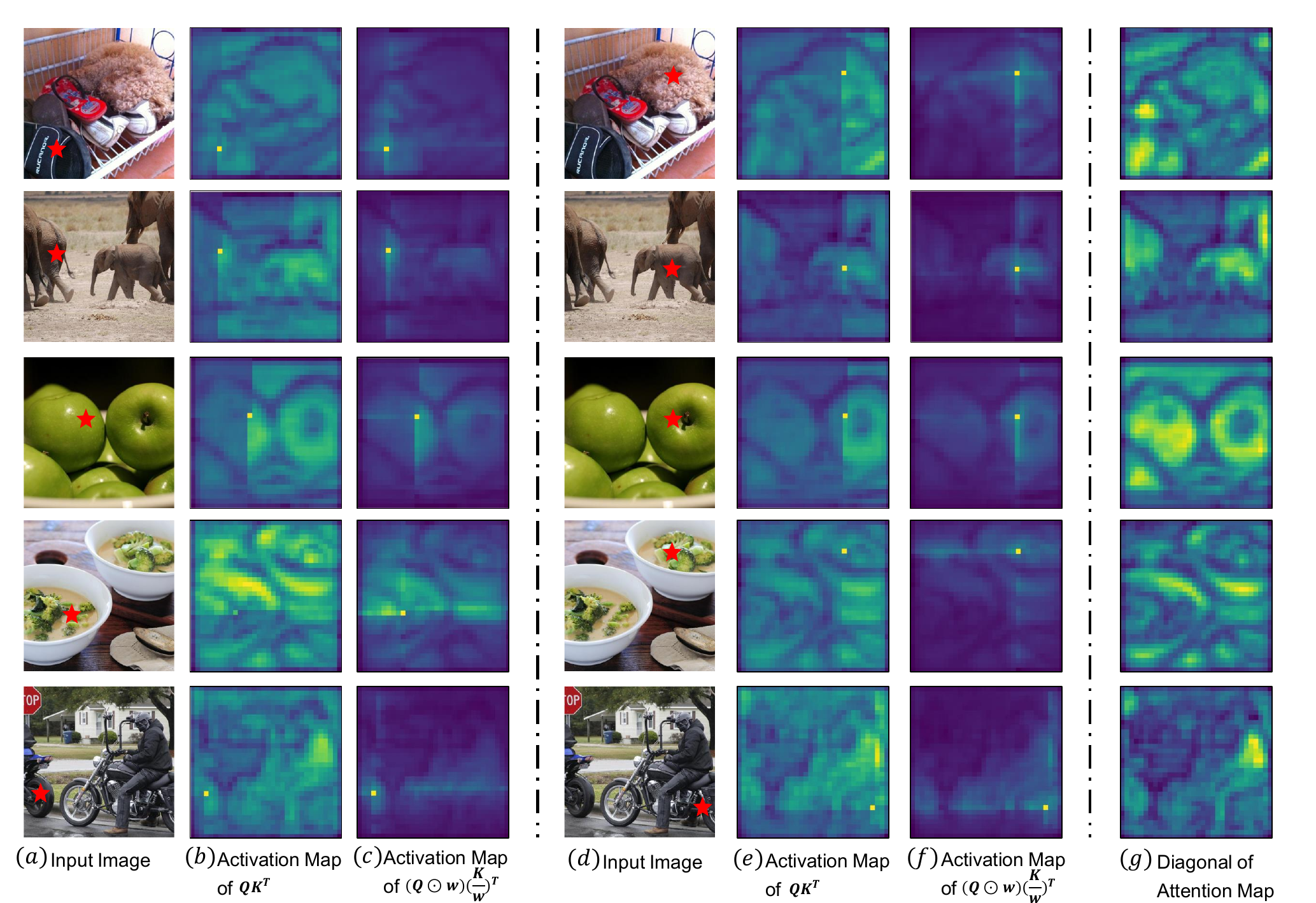}
\caption{Illustration of activation maps for the query patch (marked with a red star).}
\label{fig:app_attention_more}
\end{center}
\vskip -0.2in
\end{figure*}

Moreover, given a selected patch as the query, we visualize the corresponding activation map by reshaping the associated row in the attention map (computed by $\mathbf{Q}\mathbf{K}^\top$ or $\left(\mathbf{Q} \odot \mathbf{w}\right) \left( \mathbf{K} / \mathbf{w} \right)^\top$)
This reflects the attention score between the query patch and all previously scanned patches.
To obtain the complete visualization for a query patch, we collect and combine the activation maps from all four scanning paths in SS2D.
The visualization results of the activation map for both $\mathbf{Q}\mathbf{K}^\top$ and $\left(\mathbf{Q} \odot \mathbf{w}\right) \left( \mathbf{K} / \mathbf{w} \right)^\top$ are shown in Figure~\ref{fig:app_attention_more}.
We also visualize the diagonal elements of attention maps computed by $\left(\mathbf{Q} \odot \mathbf{w}\right) \left( \mathbf{K} / \mathbf{w} \right)^\top$, where all foreground objects are effectively highlighted and separated from the background.

\section{Detailed Experiment Settings}\label{sec:appendix_details_for_vmamba_models}

\paragraph{Network Architecture.}

The architectural specifications of Vanilla-VMamba are outlined in Table~\ref{tab:app_arch_Vanilla-VMamba}, while detailed configurations of the VMamba series are provided in Table~\ref{tab:app_arch_VMamba}.
The Vanilla-VMamba series is constructed using the vanilla VSS Block, which includes a multiplicative branch and does not have feed-forward network (FFN) layers. In contrast, the VSS Block in the VMamba series removes the multiplicative branch and introduces FFN layers.
Additionally, we provide alternative architectures for VMamba at Small and Base scales, referred to as VMamba-S[$s1l20$] and VMamba-B[$s1l20$], respectively.
The notation `$sxly$' indicates that the \texttt{ssm-ratio} is set to $x$ and the number of layers in stage 3 is set to $y$. 
Consequently, the versions presented in Table~\ref{exp:imagenet-system} can also be referred to as VMamba-S[$s2l15$] and VMamba-B[$s2l15$].

\paragraph{Experiment Setting.}

The hyper-parameters for training VMamba on ImageNet are inherited from Swin~\cite{Swin2021}, except for the parameters related to \texttt{drop\_path\_rate} and the exponential moving average (EMA) technique. 
Specifically, VMamba-T/S/B models are trained from scratch for 300 epochs, with a 20-epoch warm-up period, using a batch size of 1024.
The training process utilizes the AdamW optimizer~\cite{adamw} with betas set to $(0.9, 0.999)$, an initial learning rate of $1\times10^{-3}$, a weight decay of $0.05$, and a cosine decay learning rate scheduler. 
%
%
It is noteworthy that this is not the optimal setting for VMamba. With a learning rate of $2\times10^{-3}$, the Top-1 accuracy of VMamba-T can reach 80.7\%.

Additional techniques such as label smoothing (0.1) and EMA (decay ratio of 0.9999) are also applied.
The \texttt{drop\_path\_ratio} is set to 0.2 for Vanilla-VMamba-T and VMamba-T, 0.3 for Vanilla-VMamba-S, VMamba-S[$s2l15$] and VMamba-S[$s1l20$], 0.6 for Vanilla-VMamba-B and VMamba-B[$s2l15$], and 0.5 for VMamba-B[$s1l20$]. 
No additional training techniques are employed.

\paragraph{Throughput Evaluation.}

Detailed performance comparisons with various models are presented in Table~\ref{tab:app_imagenet_acc}.
Throughput (referred to as \texttt{TP.}) was assessed on an A100 GPU paired with an AMD EPYC 7542 CPU, utilizing the toolkit provided by~\cite{rw2019timm}.
Following the protocol outlined in~\cite{Swin2021}, we set the batch size to 128.
The training throughput (referred to as \texttt{Train TP.}) is tested on the same device with \texttt{mix-resolution}, excluding the time consumption of optimizers. 
The batch size for measuring the training throughput is also set to 128.

\paragraph{Accelerating VMamba.}

Table~\ref{tab:app_accelerate} provides detailed configurations of the intermediate variants in the acceleration process from Vanilla-VMamba-T to VMamba-T.

\paragraph{Evolution of ERF.}

We further generate the effective receptive field (ERF) maps throughout the training process for Vanilla-VMamba-T. These maps intuitively illustrate how VMamba's pattern of ERF evolves from being predominantly local to predominantly global, epoch by epoch.

\newcommand{\blocka}[3]{\multirow{6}{*}{\(\left[\begin{array}{c}
\text{\texttt{Linear} #1 $\rightarrow 2\times$#1}\\[-.1em]
\text{\texttt{DWConv} 3$\times$3, $2\times$#1}\\[-.1em]
\text{\texttt{SS2D}, dim $2\times$#1}\\[-.1em]
\text{\texttt{Linear} 2$\times$#1 $\rightarrow$ #1}\\[-.1em]
\text{\texttt{Multiplicative}}\\[-.1em]
\text{\texttt{Linear} 2$\times$#1 $\rightarrow$ #1}\\[-.1em]
\end{array}\right]\)$\times$#2}
}
\renewcommand\arraystretch{1.1}
\begin{table}[ht]
    \caption{
    Architectural overview of the Vanilla-VMamba series. Down-sampling is performed through patch merging~\cite{Swin2021} operations in stages 1, 2, and 3. The term \texttt{Linear} refers to a linear layer, while \texttt{DWConv} denotes a depth-wise convolution~\cite{han2021connection} operation. The proposed 2D-selective-scan is labeled as \texttt{SS2D}.
    }
    \label{tab:app_arch_Vanilla-VMamba}
    \begin{center}
    \resizebox{.95\linewidth}{!}{
        \begin{tabular}{c|c|c|c|c}
        \toprule
        layer name & output size & Vanilla-VMamba-T & Vanilla-VMamba-S & Vanilla-VMamba-B  \\
        \midrule
        stem & 112$\times$112 & conv 4$\times$4, 96, stride 4 & conv 4$\times$4, 96, stride 4 & conv 4$\times$4, 128, stride 4\\
        \midrule
        
        \multirow{8}{*}{stage 1} & \multirow{8}{*}{56$\times$56} & &  \\ 
        &  & vanilla VSSBLock  &  vanilla VSSBLock &  vanilla VSSBLock \\
        &  & \blocka{96}{2}{16}  & \blocka{96}{2}{16} & \blocka{128}{2}{16} \\
        &  &  &  & \\
        &  &  &  & \\
        &  &  &  & \\
        &  &  &  & \\
        &  &  &  & \\
        &  &  &  & \\
        \cline{3-5}
        &  & patch merging $\rightarrow 192$ & patch merging $\rightarrow 192$ & patch merging $\rightarrow 256$ \\
        \midrule
        
        \multirow{8}{*}{stage 2} & \multirow{8}{*}{28$\times$28} & &  \\ 
        &  & vanilla VSSBLock  &  vanilla VSSBLock &  vanilla VSSBLock \\
        &  & \blocka{192}{2}{16}  & \blocka{192}{2}{16} & \blocka{256}{2}{16} \\
        &  &  &  & \\
        &  &  &  & \\
        &  &  &  & \\
        &  &  &  & \\
        &  &  &  & \\
        &  &  &  & \\
        \cline{3-5}
        &  & patch merging $\rightarrow 384$ & patch merging $\rightarrow 384$ & patch merging $\rightarrow 512$ \\
        \midrule
        
        \multirow{8}{*}{stage 3} & \multirow{8}{*}{14$\times$14} & &  \\ 
        &  & vanilla VSSBLock  &  vanilla VSSBLock &  vanilla VSSBLock \\
        &  & \blocka{384}{9}{16}  & \blocka{384}{27}{16} & \blocka{512}{27}{16} \\
        &  &  &  & \\
        &  &  &  & \\
        &  &  &  & \\
        &  &  &  & \\
        &  &  &  & \\
        &  &  &  & \\
        \cline{3-5}
        &  & patch merging $\rightarrow 768$ & patch merging $\rightarrow 768$ & patch merging $\rightarrow 1024$ \\
        \midrule
        
        \multirow{8}{*}{stage 4} & \multirow{8}{*}{7$\times$7} & &  \\ 
        &  & vanilla VSSBLock  &  vanilla VSSBLock &  vanilla VSSBLock \\
        &  & \blocka{768}{2}{16}  & \blocka{768}{2}{16} & \blocka{1024}{2}{16} \\
        &  &  &  & \\
        &  &  &  & \\
        &  &  &  & \\
        &  &  &  & \\
        &  &  &  & \\
        &  &  &  & \\
        \midrule
        
        & 1$\times$1  & \multicolumn{3}{c}{average pool, 1000-d fc, softmax} \\
        \midrule
        \multicolumn{2}{c|}{Param. (M)} & 22.9  & 44.4  & 76.3 \\
        \midrule
        \multicolumn{2}{c|}{FLOPs} & 5.63$\times10^9$  & 11.23$\times10^9$  & 18.02$\times10^9$ \\
        \bottomrule
        \end{tabular}
    }
    \end{center}
\end{table}

\newcommand{\blockb}[3]{\multirow{5}{*}{\(\left[\begin{array}{c}
\text{\texttt{Linear} #1 $\rightarrow$ \texttt{ssm-ratio} $\times$#1}\\[-.1em]
\text{\texttt{DWConv} 3$\times$3, \texttt{ssm-ratio} $\times$#1}\\[-.1em]
\text{\texttt{SS2D}, dim \texttt{ssm-ratio} $\times$#1}\\[-.1em]
\text{\texttt{Linear} \texttt{ssm-ratio} $\times$#1 $\rightarrow$ #1}\\[-.1em]
\text{\texttt{FFN} \texttt{mlp-ratio} $\times$#1}\\[-.1em]
\end{array}\right]\)$\times$#2}
}
\begin{table}[ht]
    \caption{
    Architectural overview of the VMamba series.
    }
    \label{tab:app_arch_VMamba}
    \begin{center}
    \resizebox{.95\linewidth}{!}{
        \begin{tabular}{c|c|c|c|c|c}
        \toprule
        layer name & output size & VMamba-T & VMamba-S & VMamba-B  \\
        \midrule
        stem & 112$\times$112 & \multicolumn{3}{c}{ conv 3$\times$3 stride 2, LayerNorm, GeLU, conv 3$\times$3 stride 2, LayerNorm }\\
        \midrule
        
        \multirow{8}{*}{stage 1} & \multirow{8}{*}{56$\times$56} & &  \\ 
        &  & VSSBLock(\texttt{ssm-ratio}=1, \texttt{mlp-ratio}=4)  & VSSBlock(\texttt{ssm-ratio}=2, \texttt{mlp-ratio}=4) & VSSBLock(\texttt{ssm-ratio}=2, \texttt{mlp-ratio}=4) \\
        &  & \blockb{96}{2}{16}  & \blockb{96}{2}{16} & \blockb{128}{2}{16} \\
        &  &  &  & \\
        &  &  &  & \\
        &  &  &  & \\
        &  &  &  & \\
        &  &  &  & \\
        \cline{3-5}
        &  & \multicolumn{3}{c}{ conv 3$\times$3 stride 2, LayerNorm} \\
        \midrule
        
        \multirow{8}{*}{stage 2} & \multirow{8}{*}{28$\times$28} & &  \\ 
        &  & VSSBLock(\texttt{ssm-ratio}=1, \texttt{mlp-ratio}=4)  & VSSBlock(\texttt{ssm-ratio}=2, \texttt{mlp-ratio}=4) & VSSBLock(\texttt{ssm-ratio}=2, \texttt{mlp-ratio}=4) \\
        &  & \blockb{192}{2}{16}  & \blockb{192}{2}{16} & \blockb{256}{2}{16} \\
        &  &  &  & \\
        &  &  &  & \\
        &  &  &  & \\
        &  &  &  & \\
        &  &  &  & \\
        \cline{3-5}
        &  & \multicolumn{3}{c}{ conv 3$\times$3 stride 2, LayerNorm} \\
        \midrule
        
        \multirow{8}{*}{stage 3} & \multirow{8}{*}{14$\times$14} & &  \\ 
        &  & VSSBLock(\texttt{ssm-ratio}=1, \texttt{mlp-ratio}=4)  & VSSBlock(\texttt{ssm-ratio}=2, \texttt{mlp-ratio}=4) & VSSBLock(\texttt{ssm-ratio}=2, \texttt{mlp-ratio}=4) \\
        &  & \blockb{384}{8}{16}  & \blockb{384}{15}{16} & \blockb{512}{15}{16} \\
        &  &  &  & \\
        &  &  &  & \\
        &  &  &  & \\
        &  &  &  & \\
        &  &  &  & \\
        \cline{3-5}
        &  & \multicolumn{3}{c}{ conv 3$\times$3 stride 2, LayerNorm} \\
        \midrule
        
        \multirow{8}{*}{stage 4} & \multirow{8}{*}{7$\times$7} & &  \\
        &  & VSSBLock(\texttt{ssm-ratio}=1, \texttt{mlp-ratio}=4)  & VSSBlock(\texttt{ssm-ratio}=2, \texttt{mlp-ratio}=4) & VSSBLock(\texttt{ssm-ratio}=2, \texttt{mlp-ratio}=4) \\
        &  & \blockb{768}{2}{16}  & \blockb{768}{2}{16} & \blockb{1024}{2}{16} \\
        &  &  &  & \\
        &  &  &  & \\
        &  &  &  & \\
        &  &  &  & \\
        &  &  &  & \\
        \midrule
        
        & 1$\times$1  & \multicolumn{3}{c}{average pool, 1000-d fc, softmax} \\
        \midrule
        \multicolumn{2}{c|}{Param. (M)} & 30.2  & 50.1  & 88.6 \\
        \midrule
        \multicolumn{2}{c|}{FLOPs} & 4.91$\times10^9$  & 8.72$\times10^9$  & 15.36$\times10^9$ \\
        \bottomrule
        \end{tabular}
    }
    \end{center}
\end{table}

\begin{table}[ht]
\small
\centering
\setlength{\tabcolsep}{1pt}
\renewcommand\arraystretch{1.0}
\caption{Details of accelerating VMamba.}
\label{tab:app_accelerate}
\setlength{\tabcolsep}{0.1cm}
\resizebox{.95\linewidth}{!}{
    \begin{tabular}{l|cccc|cc|cc|cc|c}
    \Xhline{1.0pt}
    \multirow{2}{*}{Model} & \multirow{2}{*}{\texttt{d\_state}} & \multirow{2}{*}{\texttt{ssm-ratio}} & \multirow{2}{*}{\texttt{DWConv}} & \texttt{multiculative} & \texttt{layers} & \multirow{2}{*}{\texttt{FFN}} & Params & FLOPs & TP. & Train TP. & Top-1 \\
    &&&&\texttt{branch}&\texttt{numbers}&& (M) & (G) & (img/s) & (img/s) & (\%) \\
    \hline
    Vanilla-VMamba-T & 16 & 2.0  & \checkmark& \checkmark & [2,2,9,2]&& 22.9M & 5.63G  & 426 & 138 & 82.17 \\
    Step(a) & 16 & 2.0     & \checkmark& \checkmark & [2,2,9,2]&& 22.9M & 5.63G  & 467 & 165 & 82.17 \\
    Step(b) &16 & 2.0      & \checkmark& \checkmark & [2,2,9,2]&& 22.9M & 5.63G  & 464 & 184 & 82.17 \\
    Step(c) &16 & 2.0      & \checkmark& \checkmark & [2,2,9,2]&& 22.9M & 5.63G  & 638 & 195 & 82.17 \\
    \hline
    Step(d) &16 & 2.0      & & \checkmark & [2,2,2,2]&\checkmark& 29.0M & 5.63G  & 813 & 248 & 81.65 \\
    Step(d.1) &16 & 1.0      & & \checkmark & [2,2,2,2]&\checkmark& 22.9M & 4.02G  & 1336 & 405 & 81.05 \\
    Step(d.2) &16 & 1.0      & & \checkmark & [2,2,5,2]&\checkmark& 28.2M & 5.18G  & 1137 & 348 & 82.24 \\
    Step(e) &16 & 1.0      & &  & [2,2,5,2]&\checkmark& 26.2M & 4.86G  & 1179 & 360 & 82.17 \\
    Step(e.1) &16 & 1.0    & \checkmark &  & [2,2,5,2]&\checkmark& 26.3M & 4.87G  & 1164 & 358 & 82.31 \\
    Step(e.2) &1 & 1.0     & \checkmark &  & [2,2,5,2]&\checkmark& 25.6M & 3.98G  & 1942 & 647 & 81.87 \\
    Step(f) &1 & 2.0       & \checkmark&  & [2,2,5,2]&\checkmark& 30.7M & 4.86G  & 1340 & 464 & 82.49 \\
    Step(g) &1 & 1.0       & \checkmark&  & [2,2,8,2]&\checkmark& 30.2M & 4.91G  & 1686 & 571 & 82.60 \\
    \Xhline{1.0pt}
\end{tabular}
}
\end{table}

\begin{table}[ht]
\small
\centering
\setlength{\tabcolsep}{1pt}
\renewcommand\arraystretch{1.0}
\caption{Performance comparison on ImageNet-1K with an image size of 224. $^\dagger$ indicates that Vim is trained solely in float32 in practice, with a training throughput of 232.~\cite{zhu2024vision}.
}
\label{tab:app_imagenet_acc}
\setlength{\tabcolsep}{0.1cm}
    \begin{tabular}{l|ccc|cc|c}
    \Xhline{1.0pt}
    \multirow{2}{*}{Model} & Image & Params & FLOPs & TP. & Train TP. & Top-1 \\
    & Size & (M) & (G) & (img/s) & (img/s) & (\%) \\
    \hline
    DeiT-S~\cite{DeiT2021} & 224$^2$ & 22M & 4.6G  & 1761 & 2404 & 79.8 \\
    DeiT-B~\cite{DeiT2021} & 224$^2$ & 86M & 17.5G  & 503 & 1032 & 81.8 \\
    \hline
    ConvNeXt-T~\citep{liu2022convnet} & 224$^2$ & 29M & 4.5G & 1198 & 702 & 82.1 \\
    ConvNeXt-S~\citep{liu2022convnet} & 224$^2$ & 50M & 8.7G & 684 & 445 & 83.1 \\
    ConvNeXt-B~\citep{liu2022convnet} & 224$^2$ & 89M & 15.4G & 436 & 334 & 83.8 \\
    \hline
    HiViT-T~\citep{hivit} & 224$^2$ & 19M & 4.6G & 1393 & 1304 & 82.1 \\
    HiViT-S~\citep{hivit} & 224$^2$ & 38M & 9.1G & 712 & 698 & 83.5 \\
    HiViT-B~\citep{hivit} & 224$^2$ & 66M & 15.8G & 456 & 544 & 83.8 \\
    \hline
    Swin-T~\citep{Swin2021} & 224$^2$ & 28M & 4.5G & 1244 & 987 & 81.3 \\
    Swin-S~\citep{Swin2021} & 224$^2$ & 50M & 8.7G & 718 & 642 & 83.0 \\
    Swin-B~\citep{Swin2021} & 224$^2$ & 88M & 15.5G & 458 & 496 & 83.5 \\
    \hline
    XCiT-S12/16 & 224$^2$ & 26M & 4.9G & 1283 & 935 & 82.0 \\
    XCiT-S24/16 & 224$^2$ & 48M & 9.2G & 671 & 509 & 82.6 \\
    XCiT-M24/16 & 224$^2$ & 84M & 16.2G & 423 & 385 & 82.7 \\
    \hline
    S4ND-ConvNeXt-T~\cite{s4ndnguyen2022s4nd} & 224$^2$ & 30M & 5.2G & 683 & 369 & 82.2 \\
    S4ND-ViT-B~\cite{s4ndnguyen2022s4nd} & 224$^2$ & 89M & 17.1G & 398 & 400 & 80.4 \\ 
    Vim-S~\cite{zhu2024vision} & 224$^2$ & 26M & 5.3G & 811 & 344$^\dagger$ & 80.5 \\ 
    \hline
    \rowcolor{gray!20}
    Vanilla-VMamba-T & 224$^2$ & 23M & 5.6G  & 638 & 195 & 82.2 \\
    \rowcolor{gray!20}
    Vanilla-VMamba-S & 224$^2$ & 44M & 11.2G & 359 & 111 & 83.5 \\
    \rowcolor{gray!20}
    Vanilla-VMamba-B & 224$^2$ & 76M & 18.0G & 268 & 84 & 83.7 \\
    \hline
    \rowcolor{gray!20}
    VMamba-T & 224$^2$ & 30M & 4.9G  & 1686 & 571 & 82.6 \\
    \rowcolor{gray!20}
    VMamba-S[$s2l15$] & 224$^2$ & 50M & 8.7G & 877 & 314 & 83.6 \\
    \rowcolor{gray!20}
    VMamba-B[$s2l15$] & 224$^2$ & 89M & 15.4G & 646 & 247 & 83.9 \\
    \hline
    \rowcolor{gray!20}
    VMamba-S[$s1l20$] & 224$^2$ & 49M & 8.6G & 1106 & 390 & 83.3 \\
    \rowcolor{gray!20}
    VMamba-B[$s1l20$] & 224$^2$ & 87M & 15.2G & 827 & 313 & 83.8 \\
    \Xhline{1.0pt}
\end{tabular}
\end{table}

\begin{figure*}
    \centering
    \vskip 0.2in
    \begin{center}
    \includegraphics[width=0.99\textwidth]{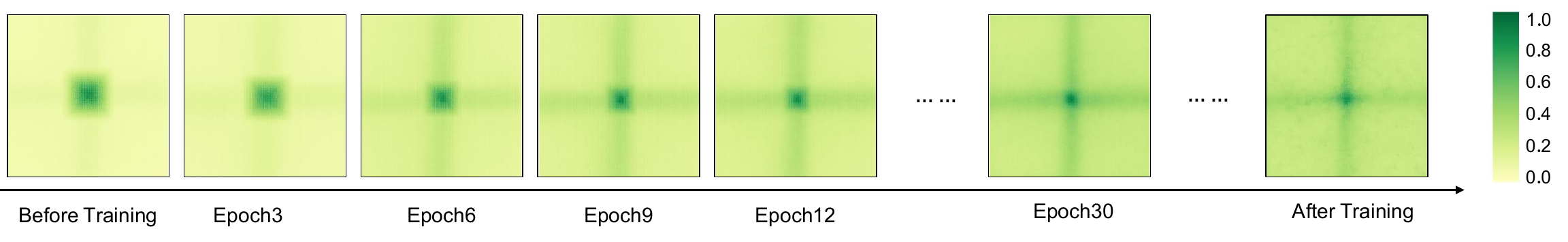}
    \caption{Illustration of ERF maps throughout the training process of Vanilla-VMamba-T (with EMA).}
    \label{fig:erf_training}
    \end{center}
    \vskip -0.2in
\end{figure*}

\section{Performance of the VMamba Family on Downstream Tasks}\label{sec:details_for_vmamba_on_downstream_tasks}

In this section, we present the experimental results of Vanilla-VMamba and VMamba on the MSCOCO and ADE20k datasets. 
The results are summarized in Table~\ref{tab:app_coco} and Table~\ref{tab:app_ade20k}, respectively.

For object detection and instance segmentation, we adhere to the protocol outlined by Swin~\cite{Swin2021} and construct our models using the mmdetection framework~\cite{MMdet2019}.
Specifically, we utilize the AdamW optimizer~\cite{adamw} and fine-tune the classification models pre-trained on ImageNet-1K for both 12 and 36 epochs. 
The learning rate is initialized at $1\times 10^{-4}$ and decreased by a factor of 10 at the 9-\textit{th} and 11-\textit{th} epoch. 
We incorporate multi-scale training and random flipping with a batch size of 16, following established practices for object detection evaluations.

For semantic segmentation, we follow Swin~\cite{Swin2021} and construct a UperHead~\cite{upernet} network on top of the pre-trained model using the MMSegmentation library~\cite{mmseg2020}.
We employ the AdamW optimizer~\cite{adamw} and set the learning rate to $6\times10^{-5}$. 
The fine-tuning process spans a total of $160k$ iterations with a batch size of 16. 
The default input resolution is $512\times512$.

\begin{table}[ht]
\caption{Object detection and instance segmentation results on COCO dataset. FLOPs are calculated using inputs of size $1280\times800$. Here, $AP^{b}$ and $AP^{m}$ denote box AP and mask AP, respectively. "$1\times$" indicates models fine-tuned for 12 epochs, while "$3\times$MS" signifies the utilization of multi-scale training for $36$ epochs.
}
\label{tab:app_coco}
\centering
\setlength{\tabcolsep}{0.2cm}
\begin{tabular}{l|ccc|ccc|cc}
    \toprule
    \multicolumn{9}{c}{\textbf{Mask R-CNN 1$\times$ schedule}}\\
    \midrule
    Backbone & AP$^\text{b}$ & AP$^\text{b}_\text{50}$ & AP$^\text{b}_\text{75}$ & AP$^\text{m}$ & AP$^\text{m}_\text{50}$ & AP$^\text{m}_\text{75}$ & Params & FLOPs \\
    \midrule
    Swin-T & 42.7 & 65.2 & 46.8 & 39.3 & 62.2 & 42.2 & 48M & 267G \\
    ConvNeXt-T & 44.2 & 66.6 & 48.3 & 40.1 & 63.3 & 42.8 & 48M & 262G \\
    \rowcolor{gray!20}
    Vanilla-VMamba-T & 46.5 & 68.5 & 50.7 & 42.1 & 65.5 & 45.3 & 42M & 286G \\
    \rowcolor{gray!20}
    VMamba-T & 47.3 & 69.3 & 52.0 & 42.7 & 66.4 & 45.9 & 50M & 271G \\
    \midrule
    Swin-S & 44.8 & 66.6 & 48.9 & 40.9 & 63.4 & 44.2 & 69M & 354G \\
    ConvNeXt-S & 45.4 & 67.9 & 50.0 & 41.8 & 65.2 & 45.1 & 70M & 348G \\
    \rowcolor{gray!20}
    Vanilla-VMamba-S & 48.2 & 69.7 & 52.5 & 43.0 & 66.6 & 46.4 & 64M & 400G \\
    \rowcolor{gray!20}
    VMamba-S & 48.7 & 70.0 & 53.4 & 43.7 & 67.3 & 47.0 & 70M & 349G \\
    \midrule
    Swin-B & 46.9 & -- & -- & 42.3 & -- & -- & 107M & 496G \\
    ConvNeXt-B & 47.0 & 69.4 & 51.7 & 42.7 & 66.3 & 46.0 & 108M & 486G \\
    \rowcolor{gray!20}
    Vanilla-VMamba-B & 48.6 & 70.0 & 53.1 & 43.3 & 67.1 & 46.7 & 96M & 540G \\
    \rowcolor{gray!20}
    VMamba-B & 49.2 & 71.4 & 54.0 & 44.1 & 68.3 & 47.7 & 108M & 485G \\
    \midrule
    \multicolumn{9}{c}{\textbf{Mask R-CNN 3$\times$ MS schedule}}\\
    \midrule
    Swin-T &  46.0 & 68.1 & 50.3 & 41.6 & 65.1 & 44.9 & 48M & 267G \\
    ConvNeXt-T &  46.2 & 67.9 & 50.8 & 41.7 & 65.0 & 44.9 & 48M & 262G \\
    \rowcolor{gray!20}
    Vanilla-VMamba-T & 48.5 & 70.0 & 52.7 & 43.2 & 66.9 & 46.4 & 42M & 286G \\
    \rowcolor{gray!20}
    VMamba-T & 48.8 & 70.4 & 53.5 & 43.7 & 67.4 & 47.0 & 50M & 271G \\
    \midrule
    Swin-S & 48.2 & 69.8 & 52.8 & 43.2 & 67.0 & 46.1 & 69M & 354G \\
    ConvNeXt-S & 47.9 & 70.0 & 52.7 & 42.9 & 66.9 & 46.2 & 70M & 348G \\
    \rowcolor{gray!20}
    Vanilla-VMamba-S & 49.7 & 70.4 & 54.2 & 44.0 & 67.6 & 47.3 & 64M & 400G \\
    \rowcolor{gray!20}
    VMamba-S & 49.9 & 70.9 & 54.7 & 44.2 & 68.2 & 47.7 & 70M & 349G \\
\bottomrule
\end{tabular}
\end{table}

\begin{table}[ht]
\caption{Semantic segmentation results on ADE20K using UperNet~\cite{upernet}.
We evaluate the performance of semantic segmentation on the ADE20K dataset with UperNet~\cite{upernet}. FLOPs are calculated with input sizes of $512\times2048$. "SS" and "MS" denote single-scale and multi-scale testing, respectively.
}
\label{tab:app_ade20k}
\centering
\setlength{\tabcolsep}{0.2cm}
\begin{tabular}{c|c|cc|ccc}
    \toprule
    method & crop size & mIoU (SS) & mIoU (MS) & Params & FLOPs \\
    \midrule
    Swin-T & $512^{2}$ & 44.5 & 45.8 & 60M & 945G \\
    ConvNeXt-T & $512^{2}$ & 46.0 & 46.7 & 60M & 939G \\
    \rowcolor{gray!20}
    Vanilla-VMamba-T & $512^{2}$ & 47.3 & 48.3 & 55M & 964G  \\
    \rowcolor{gray!20}
    VMamba-T & $512^{2}$ & 48.0 & 48.8 & 62M & 949G  \\
    \midrule
    Swin-S & $512^{2}$ & 47.6 & 49.5 & 81M & 1039G \\
    ConvNeXt-S & $512^{2}$ & 48.7 & 49.6 & 82M & 1027G \\
    \rowcolor{gray!20}
    Vanilla-VMamba-S & $512^{2}$ & 49.5 & 50.5 & 76M & 1081G  \\
    \rowcolor{gray!20}
    VMamba-S & $512^{2}$ & 50.6 & 51.2 & 82M & 1028G  \\
    \midrule
    Swin-B & $512^{2}$ & 48.1 & 49.7 & 121M & 1188G \\
    ConvNeXt-B & $512^{2}$ & 49.1 & 49.9 & 122M & 1170G \\
    \rowcolor{gray!20}
    Vanilla-VMamba-B & $512^{2}$ & 50.0 & 51.3 & 110M & 1226G  \\
    \rowcolor{gray!20}
    VMamba-B & $512^{2}$ & 51.0 & 51.6 & 122M & 1170G  \\
    \bottomrule
\end{tabular}
\end{table}

\section{Details of VMamba's Scale-Up Experiments}\label{sec:appendix_details_for_vmamba_scale_up}
Given Mamba's exceptional ability in efficient long sequence modeling, we conduct experiments to assess whether VMamba inherits this characteristic. We evaluate the computational efficiency and classification accuracy of VMamba with progressively larger input spatial resolutions.
Specifically, following the protocol in XCiT~\cite{xcit}, we apply VMamba, trained on $224\times224$ inputs, to images with resolutions ranging from $288\times288$ to $768\times768$. 
We measure the generalization performance in terms of the number of parameters, FLOPs, throughput during both training and inference, and the top-1 classification accuracy on ImageNet-1K.
We also conduct experiments under the `\texttt{linear tuning}' setting, where only the header network, consisting of a single linear module, is fine-tuned from random initialization using features extracted by the backbone models.

According to the results summarized in Table~\ref{tab:input_scaling}, VMamba demonstrates the most stable performance across (\textit{i.e.}, modest performance drop) different input image sizes, achieving a top-1 classification accuracy of $74.7\%$ without fine-tuning ($79.2\%$ with \texttt{linear tuning}), while maintaining a relatively high throughput of 149 images per second at an input resolution of $768\times768$.
In comparison, Swin~\cite{Swin2021} achieves the second-highest performance with a top-1 accuracy of $73.1\%$ without fine-tuning ($77.5\%$ under \texttt{linear tuning}) at the same input size, using scaled window sizes (set as the resolution divided by $32$).
However, its throughput significantly drops to $53$ images per second.
Furthermore, ConvNeXt~\cite{liu2022convnet} maintains a relatively high inference speed (\textit{i.e.}, a throughput of $103$ images per second) at the largest input resolution. 
However, its classification accuracy drops to $69.5\%$ when directly tested on images of size $768\times768$, indicating its limited adaptability to images with large spatial resolutions.
Deit-S also shows a dramatic performance drop, primarily due to the interpolation used in the absolute positional embedding.

Notably, VMamba displays a linear increase in computational complexity, as measured by FLOPs, which is comparable to CNN-based architectures. 
This finding aligns with the theoretical conclusions drawn from selective SSMs~\cite{mambagu2023mamba}.

\begin{table}[ht]
    \vspace{-4pt}
    \centering
    \caption{Comparison of generalizability to inputs with increased spatial resolutions. The throughput and training throughput are measured with a batch size of $32$ using PyTorch 2.0 on an A100 GPU paired with an AMD EPYC 7542 CPU. Unlike throughput, the model's forward pass, loss calculation, and backward pass are included in calculating the training throughput, with mixed precision. 
    We re-implemented the HiViT-T, as the checkpoint for HiViT-T has not been released. 
    ${\dagger}$ denotes that the batch size $\le$ 16 due to out-of-memory (OOM) issues. 
    }
    \label{tab:input_scaling}
    \setlength{\tabcolsep}{0.2cm}
    \resizebox{0.9\linewidth}{!}{
        \ra{1.1}
        \begin{tabular}{c|ccccc|cc}
        \toprule
        \multirow{2}{*}{Model} 
        &  Image & Param. & FLOPs & TP. & Train TP. & Top-1 & Top-1 acc. (\%) \\
        &  Size & (M) & (G) & (img/s) & (img/s) & acc. (\%)& (w/ \texttt{linear tuning}) \\
        \midrule
        \multicolumn{8}{c}{\textbf{SSM-Based}} \\
        \midrule
        \multirow{7}{*}{VMamba-Tiny}
        & 224$^2$ & 30M & 4.91G & 1490 & 418 & 82.60 & 82.64 \\
        & 288$^2$ & 30M & 8.11G & 947 & 303 & 82.95 & 83.03 \\
        & 384$^2$ & 30M & 14.41G & 566 & 187 & 82.41 & 82.77 \\
        & 512$^2$ & 30M & 25.63G & 340 & 121 & 80.92 & 81.88 \\
        & 640$^2$ & 30M & 40.04G & 214 & 75 & 78.60 & 80.62 \\
        & 768$^2$ & 30M & 57.66G & 149 & 53 & 74.66 & 79.22 \\
        \midrule
        \multirow{7}{*}{VMamba-Tiny[$s2l5$]}
        & 224$^2$ & 31M & 4.86G & 1227 & 399 & 82.49 & 82.52 \\
        & 288$^2$ & 31M & 8.03G & 761 & 255 & 82.81 & 82.93 \\
        & 384$^2$ & 31M & 14.27G & 452 & 155 & 82.51 & 82.74 \\
        & 512$^2$ & 31M & 25.38G & 272 & 100 & 81.07 & 82.02 \\
        & 640$^2$ & 31M & 39.65G & 170 & 60 & 79.30 & 81.02 \\
        & 768$^2$ & 31M & 57.09G & 117 & 42 & 76.06 & 79.69 \\
        \midrule
        \multirow{7}{*}{Vanilla-VMamba-Tiny}
        & 224$^2$ & 23M & 5.63G & 628 & 189 & 82.17 & 82.09 \\
        & 288$^2$ & 23M & 9.30G & 390 & 117 & 82.74 & 82.76 \\
        & 384$^2$ & 23M & 16.53G & 212 & 65 & 82.40 & 82.72 \\
        & 512$^2$ & 23M & 29.39G & 138 & 53 & 81.05 & 81.97 \\
        & 640$^2$ & 23M & 45.93G & 87 & 27 & 78.79 & 80.71 \\
        & 768$^2$ & 23M & 66.14G & 52 & 18 & 75.09 & 79.12 \\
        \midrule
        \multicolumn{8}{c}{\textbf{Transformer-Based}} \\
        \midrule
        \multirow{7}{*}{Swin-Tiny}
        & 224$^2$ & 28M & 4.51G & 1142 & 769 & 81.19 & 81.18 \\
        & 288$^2$ & 28M & 7.60G & 638 & 489 & 81.46 & 81.62 \\
        & 384$^2$ & 28M & 14.05G & 316 & 268 & 80.67 & 81.12 \\
        & 512$^2$ & 28M & 26.65G & 176 & 131 & 78.97 & 80.21 \\
        & 640$^2$ & 28M & 45.00G & 88 & 68 & 76.55 & 78.89 \\
        & 768$^2$ & 29M & 70.72G & 53 & 38 & 73.06 & 77.54 \\
        \midrule
        \multirow{7}{*}{XCiT-S12/16}
        & 224$^2$ & 26M & 4.87G & 1127 & 505 & 81.87 & 81.89 \\
        & 288$^2$ & 26M & 8.05G & 724 & 462 & 82.44 & 82.44 \\
        & 384$^2$ & 26M & 14.31G & 425 & 308 & 81.84 & 82.21 \\
        & 512$^2$ & 26M & 25.44G & 244 & 185 & 79.80 & 80.92 \\
        & 640$^2$ & 26M & 39.75G & 158 & 122 & 76.84 & 79.00 \\
        & 768$^2$ & 26M & 57.24G & 111 & 87 & 72.52 & 76.92 \\
        \midrule
        \multirow{7}{*}{HiViT-Tiny}
        & 224$^2$ & 19M & 4.60G & 1261 & 1041 & 81.92 & 81.85 \\
        & 288$^2$ & 19M & 7.93G & 750 & 614 & 82.45 & 82.42 \\
        & 384$^2$ & 19M & 15.21G & 388 & 333 & 81.51 & 81.91 \\
        & 512$^2$ & 20M & 30.56G & 186 & 150 & 79.30 & 80.49 \\
        & 640$^2$ & 20M & 54.83G & 93 & 71 & 76.09 & 78.58 \\
        & 768$^2$ & 20M & 91.41G & 55 & 37$^\dagger$ & 71.38 & 76.47 \\
        \midrule
        \multirow{7}{*}{DeiT-Small}
        & 224$^2$ & 22M & 4.61G & 1573 & 1306 & 80.69 & 80.40 \\
        & 288$^2$ & 22M & 7.99G & 914 & 1124 & 80.80 & 80.63 \\
        & 384$^2$ & 22M & 15.52G & 502 & 697 & 78.87 & 79.54 \\
        & 512$^2$ & 22M & 31.80G & 261 & 387 & 74.21 & 76.91 \\
        & 640$^2$ & 23M & 58.17G & 149 & 244 & 68.04 & 73.31 \\
        & 768$^2$ & 23M & 98.70G & 90 & 156 & 60.98 & 69.62 \\
        \midrule
        \multicolumn{8}{c}{\textbf{ConvNet-Based}} \\
        \midrule
        \multirow{7}{*}{ConvNeXt-Tiny}
        & 224$^2$ & 29M & 4.47G & 1107 & 614 & 82.05 & 81.95 \\
        & 288$^2$ & 29M & 7.38G & 696 & 403 & 82.23 & 82.30 \\
        & 384$^2$ & 29M & 13.12G & 402 & 240 & 81.05 & 81.78 \\
        & 512$^2$ & 29M & 23.33G & 226 & 140 & 78.03 & 80.37 \\
        & 640$^2$ & 29M & 36.45G & 147 & 90 & 74.27 & 78.77 \\
        & 768$^2$ & 29M & 52.49G & 103 & 63 & 69.50 & 76.89 \\
        \bottomrule
        \end{tabular}
    }
\end{table}

\section{Ablation Study}\label{sec:appendix_ablations}

\subsection{Influence of the Scanning Pattern}\label{sec:apdx_scan_methods}

\begin{figure*}
    \centering
    \vskip 0.2in
    \begin{center}
    \includegraphics[width=0.8\textwidth]{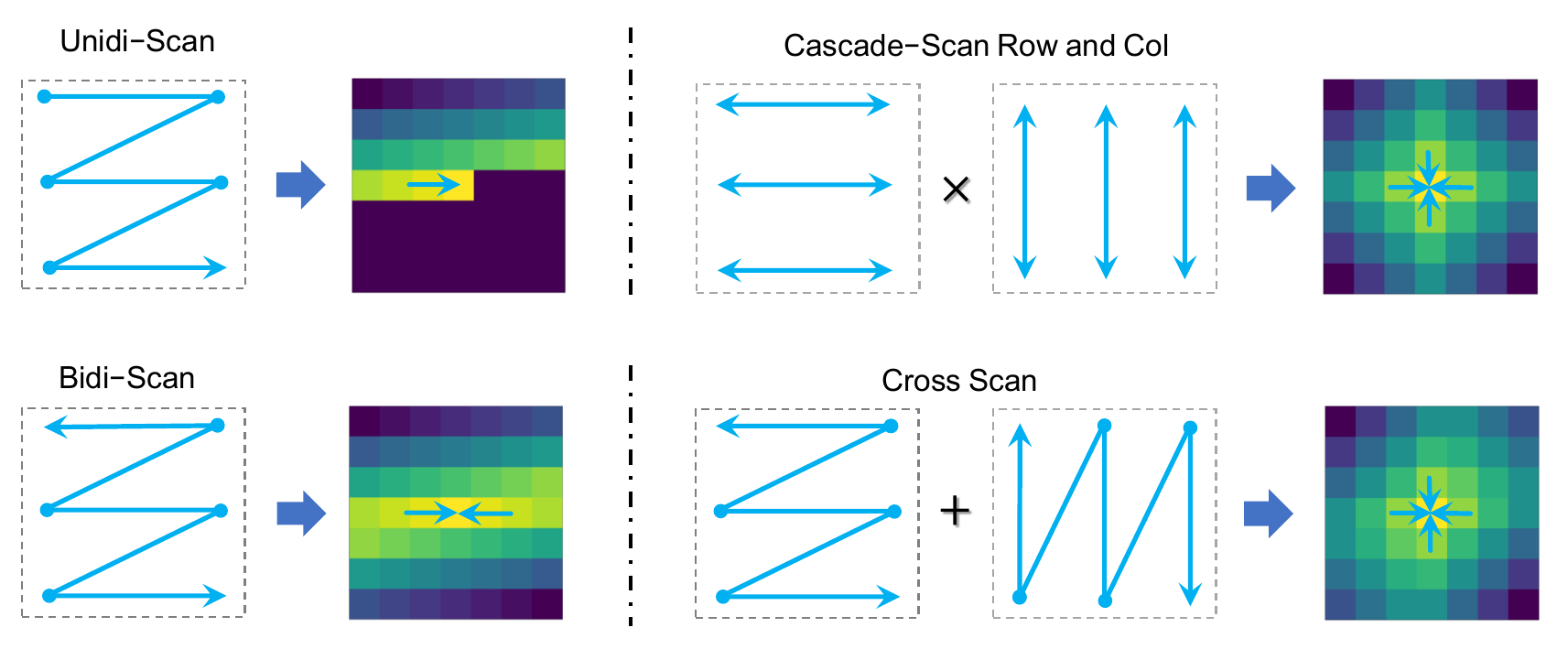}
    \caption{Illustration of different scanning patterns for selective scan.
    }
    \label{fig:scans_approaches}
    \end{center}
    \vskip -0.2in
\end{figure*}

In the main submission, we validate the effectiveness of the proposed scanning pattern (referred to as Cross-Scan) in SS2D by comparing it to three alternative image traversal approaches, \textit{i.e.}, Unidi-Scan, Bidi-Scan, and Cascade-Scan (Figure~\ref{fig:scans_approaches}).
Notably, since Unidi-Scan, Bidi-Scan, and Cross-Scan are all implemented in \texttt{Triton}, they exhibit minimal differences in throughput.
The results in Table~\ref{tab:app_scan_approaches} indicate that Cross-Scan demonstrates superior data modeling capacity, as reflected by its higher classification accuracy. This advantage likely stems from the two-dimensional prior introduced by the four-way scanning design.
Nevertheless, the practical implementation of Cascade-Scan is significantly constrained by its relatively slow computational pace, primarily due to the inadequate compatibility between selective scanning and high-dimensional data, which is further affected by the multi-step scanning procedure.

\begin{figure*}
    \centering
    \vskip 0.2in
    \begin{center}
    \includegraphics[width=0.99\textwidth]{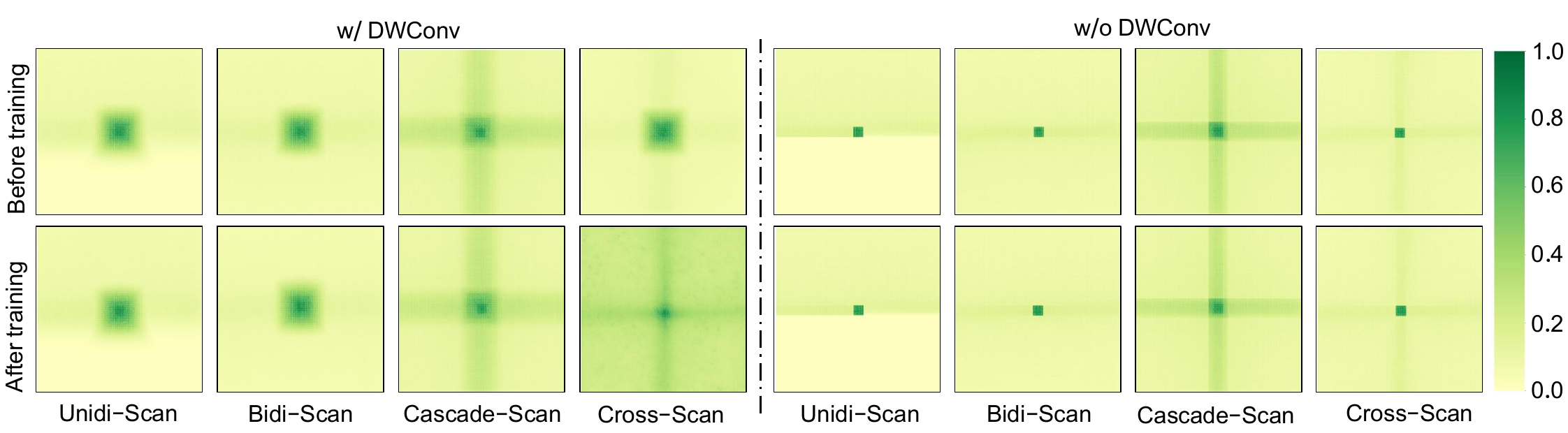}
    \caption{The visualization of ERF for models with different scanning patterns.}
    \label{fig:erf_scanmethod}
    \end{center}
    \vskip -0.2in
\end{figure*}

Figure~\ref{fig:erf_scanmethod} indirectly demonstrates that among the analyzed scanning methods, only Bidi-Scan, Cascade-Scan, and Cross-Scan showcase global ERFs. 
Moreover, only Cross-Scan and Cascade-Scan exhibit two-dimensional (2D) priors. 
It is also worth noting that \texttt{DWConv}~\cite{han2021connection} plays a critical role in establishing 2D priors, thereby contributing to the formation of global ERFs.

\begin{table}[ht]
    \small
    \centering
    \setlength{\tabcolsep}{0.2cm}
    \caption{The performance of VMamba-T with different scanning patterns.
    }
    \label{tab:app_scan_approaches}
    \setlength{\tabcolsep}{0.1cm}
    \begin{tabular}{l|cc|cc|c}
    \Xhline{1.0pt}
    \multirow{2}{*}{Model} & Params & FLOPs & TP. & Train TP. & Top-1 \\
     & (M) & (G) & (img/s) & (img/s) & (\%) \\
    \hline
    \multicolumn{6}{c}{\textbf{VMamba w/ dwconv}}\\
    \hline
    Unidi-Scan & 30.2M & 4.91G  & 1682 & 571 & 82.26 \\
    Bidi-Scan & 30.2M & 4.91G & 1687 & 572 & 82.49 \\
    Cascade-Scan & 30.2M & 4.91G  & 998 & 308 & 82.42 \\
    Cross-Scan & 30.2M & 4.91G & 1686 & 571 & 82.60 \\
    \hline
    \multicolumn{6}{c}{\textbf{VMamba w/o dwconv}}\\
    \hline
    Unidi-Scan & 30.2M & 4.89G & 1716 & 578 & 80.88 \\
    Bidi-Scan & 30.2M & 4.89G & 1719 & 578 & 81.80 \\
    Cascade-Scan & 30.2M & 4.90G  & 1007 & 309 & 81.71 \\
    Cross-Scan & 30.2M & 4.89G & 1717 & 577 & 82.25 \\
    \Xhline{1.0pt}
\end{tabular}
\end{table}

\subsection{Influence of the Initialization Approach}\label{sec:apdx_init}

In our study, we adopted the initialization scheme originally proposed for the SS2D block in S4D~\cite{s4dgu2022}. 
Therefore, it is necessary to investigate the contribution of this initialization method to the effectiveness of VMamba. 
To explore this further, we replaced the default initialization with two alternative methods: random initialization and zero initialization.

For both initialization methods, we set the parameter $\mathbf{D}$ in equation~\ref{eq:ssm} to a vector of all ones, mimicking a basic skip connection (thus we have $\mathbf{y} = \mathbf{C} \mathbf{h} + \mathbf{D} \mathbf{u}$). 
Additionally, the weights and biases associated with the transformation to the dimension $D_v$ (which matches the input size), are initialized as random vectors. In contrast, Mamba~\cite{mambagu2023mamba} employs a more sophisticated initialization.

The main distinction between random and zero initialization lies in the parameter $A$ in equation~\ref{eq:ode_dis_app}, which is typically initialized as a HiPPO matrix in both Mamba~\cite{mambagu2023mamba,s4dgu2022} and our implementation of VMamba. 
Given that we selected the hyper-parameter \texttt{d\_state} to be 1, the Mamba initialization for $log(A)$ can be simplified to all zeros, which aligns with zero initialization. 
In contrast, random initialization assigns a random vector to $log(A)$.
We choose to initialize $log(A)$ rather than $A$ directly to keep $A$ near the all-ones matrix when the network parameters are close to zero, which empirically enhances the training stability.

\begin{figure*}
    \centering
    \vskip 0.2in
    \begin{center}
    \includegraphics[width=0.5\textwidth]{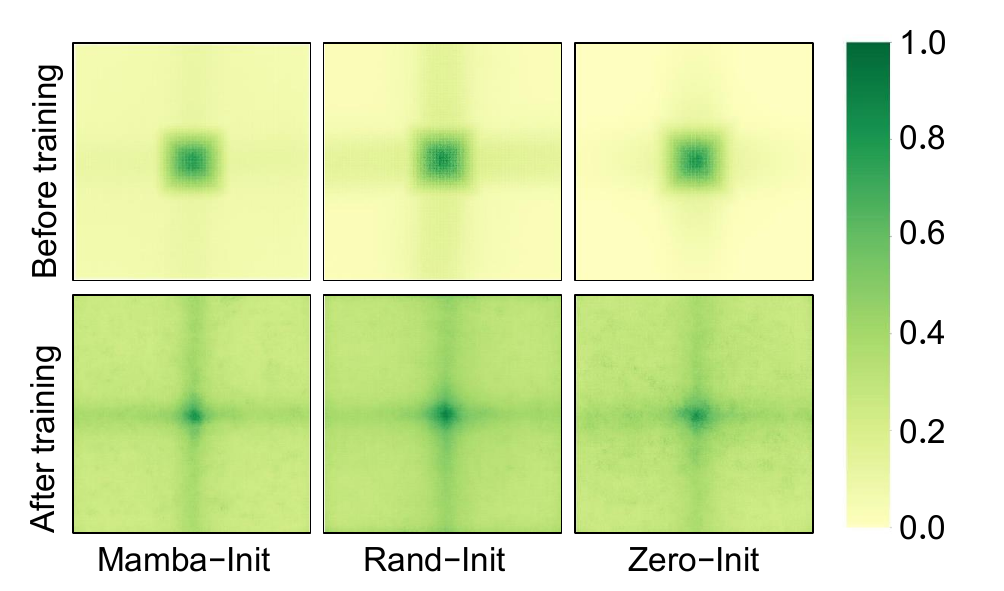}
    \caption{The visualization of ERF of VMamba with different initialization.}
    \label{fig:erf_init}
    \end{center}
    \vskip -0.2in
\end{figure*}

The experimental results in Table~\ref{tab:app_initialization} indicate that, at least for image classification with SS2D blocks, the model's performance is not significantly affected by the initialization method.
%
%
Therefore, within this context, the sophisticated initialization method employed in Mamba~\cite{mambagu2023mamba} can be substituted with a simpler, more straightforward approach.
We also visualize the ERF maps of models trained with different initialization methods (see Figure~\ref{fig:erf_init}), which intuitively reflect SS2D's robustness across various initialization schemes.

\begin{table}[ht]
    \small
    \centering
    \setlength{\tabcolsep}{1pt}
    \renewcommand\arraystretch{1.0}
    \caption{The performance of VMamba-T with different initialization.
    }
    \label{tab:app_initialization}
    \setlength{\tabcolsep}{0.1cm}
    \begin{tabular}{c|cc|cc|c}
    \Xhline{1.0pt}
    \multirow{2}{*}{\texttt{initialization}} & Params & FLOPs & TP. & Train TP. & Top-1 \\
    & (M) & (G) & (img/s) & (img/s) & acc. (\%) \\
    \hline
    mamba & 30.2 & 4.91 & 1686 & 571 & 82.60 \\
    rand & 30.2 & 4.91 & 1682 & 570 & 82.58 \\
    zero & 30.2 & 4.91 & 1683 & 570 & 82.67 \\
    \Xhline{1.0pt}
\end{tabular}
\end{table}

\subsection{Influence of the \texttt{d\_state} Parameter}\label{sec:apdx_dstate}

Throughout this study, we primarily set the value of \texttt{d\_state} to 1 to optimize VMamba's computational speed.
To further explore the impact of \texttt{d\_state} on the model's performance, we conduct a series of experiments.

As shown in Table~\ref{tab:app_dstate}, with all other hyper-parameters fixed, we increase \texttt{d\_state} from 1 to 4. 
This results in a slight improvement in performance but a substantial decrease in throughput, indicating a significant negative impact on the VMamba's computational efficiency.
However, increasing \texttt{d\_state} to 8, while reducing \texttt{ssm-ratio} to maintain computational complexity, leads to improved accuracy. 
Moreover, when \texttt{d\_state} is further increased to 16, with \texttt{ssm-ratio} set to 1, performance declines.

These findings suggest that modest increases in \texttt{d\_state} may not necessarily lead to better performance. 
Instead, selecting the optimal combination of \texttt{d\_state} and \texttt{ssm-ratio} is crucial for achieving a good trade-off between inference speed and performance.

\begin{table}[ht]
    \small
    \centering
    \setlength{\tabcolsep}{1pt}
    \renewcommand\arraystretch{1.0}
    \caption{The performance of VMamba-T with different \texttt{d\_state}.
    }
    \label{tab:app_dstate}
    \setlength{\tabcolsep}{0.1cm}
    \begin{tabular}{cc|cc|cc|c}
    \Xhline{1.0pt}
    \multirow{2}{*}{\texttt{d\_state}} & \multirow{2}{*}{\texttt{ssm-ratio}} & Params & FLOPs & TP. & Train TP. & Top-1\\
    && (M) & (G) & (img/s) & (img/s) & acc. (\%) \\
    \hline
    1 & 2.0  & 30.7 & 4.86 & 1340 & 464 & 82.49 \\
    2 & 2.0  & 30.8 & 4.98 & 1269 & 432 & 82.50 \\
    4 & 2.0  & 31.0 & 5.22 & 1147 & 382 & 82.51 \\
    8 & 1.5  & 28.6 & 5.04 & 1148 & 365 & 82.69 \\
    16 & 1.0 & 26.3 & 4.87 & 1164 & 358 & 82.31 \\
    \Xhline{1.0pt}
\end{tabular}
\end{table}

\subsection{Influence of \texttt{ssm-ratio}, \texttt{mlp-ratio}, and \texttt{layer numbers}}\label{sec:apdx_sratio}

In this section, we investigate the trade-offs among \texttt{ssm-ratio}, \texttt{layer numbers}, and \texttt{mlp-ratio}.

Experimental results shown in Table~\ref{tab:app_ssmratio} indicate that reducing \texttt{ssm-ratio} significantly decreases performance but substantially improves inference speed.
Conversely, increasing \texttt{layer numbers} enhances the performance while slowing down the model.

As the hyper-parameter \texttt{ssm-ratio} represents the dimension used by the SS2D module, the trade-off between \texttt{ssm-ratio} and \texttt{layer numbers} can be interpreted as a balance between \texttt{channel-mixing} and \texttt{token-mixing}~\cite{tolstikhin2021mlp}.
Furthermore, we reduce \texttt{mlp-ratio} from 4.0 to 2.0 and progressively increase \texttt{ssm-ratio} to maintain constant FLOPs, as shown in Table~\ref{tab:app_mlpratio}.
The results presented in Tables \ref{tab:app_ssmratio} and \ref{tab:app_mlpratio} highlight the importance of an optimal combination of \texttt{ssm-ratio}, \texttt{mlp-ratio}, and \texttt{layer numbers} for constructing a model that balances effectiveness and efficiency.

\begin{table}[ht]
    \small
    \centering
    \setlength{\tabcolsep}{1pt}
    \renewcommand\arraystretch{1.0}
    \caption{The performance of VMamba-T under different combination of  \texttt{ssm-ratio} and \texttt{layer numbers}.
    }
    \label{tab:app_ssmratio}
    \setlength{\tabcolsep}{0.1cm}
    \begin{tabular}{cc|cc|cc|c}
    \Xhline{1.0pt}
    \multirow{2}{*}{\texttt{ssm-ratio}} & \texttt{layer} & Params & FLOPs & TP. & Train TP. & Top-1 \\
    & \texttt{numbers} & (M) & (G) & (img/s) & (img/s) & acc. (\%) \\
    \hline
    2.0 & [2,2,5,2] & 30.7 & 4.86 & 1340 & 464 & 82.49 \\
    1.0 & [2,2,5,2] & 25.6 & 3.98 & 1942 & 647 & 81.87 \\
    1.0 & [2,2,8,2] & 30.2 & 4.91 & 1686 & 571 & 82.60 \\
    \Xhline{1.0pt}
\end{tabular}
\end{table}

\begin{table}[ht]
    \small
    \centering
    \setlength{\tabcolsep}{1pt}
    \renewcommand\arraystretch{1.0}
    \caption{The performance of VMamba under different combination of  \texttt{ssm-ratio} and \texttt{mlp-ratio}.
    }
    \label{tab:app_mlpratio}
    \setlength{\tabcolsep}{0.1cm}
    \begin{tabular}{cc|cc|cc|c}
    \Xhline{1.0pt}
     \multirow{2}{*}{\texttt{mlp-ratio}} & \multirow{2}{*}{\texttt{ssm-ratio}} & Params & FLOPs & TP. & Train TP. & Top-1 \\
    && (M) & (G) & (img/s) & (img/s) & acc. (\%) \\
    \hline
    4.0 & 1.0 & 30.2 & 4.91 & 1686 & 571 & 82.60 \\
    3.0 & 1.5 & 28.5 & 4.65 & 1419 & 473 & 82.75 \\
    2.0 & 2.5 & 29.9 & 4.95 & 1075 & 361 & 82.86 \\
    \Xhline{1.0pt}
\end{tabular}
\end{table}

\subsection{Influence of the Activation Function}\label{sec:apdx_activation}

In VMamba, the SiLU~\cite{elfwing2018sigmoid} activation function is utilized to build the SS2D block. 
However, experimental results in Table~\ref{tab:app_activation} show that VMamba maintains robustness across different activation functions.
This implies that the choice of activation function does not substantially affect the model's performance. 
Therefore, there is flexibility to choose an appropriate activation function based on computational constraints or other preferences.

\begin{table}[ht]
    \small
    \centering
    \setlength{\tabcolsep}{1pt}
    \renewcommand\arraystretch{1.0}
    \caption{The performance of VMamba-T with different activation functions in SS2D.
    }
    \label{tab:app_activation}
    \setlength{\tabcolsep}{0.1cm}
    \begin{tabular}{c|cc|cc|c}
    \Xhline{1.0pt}
    \multirow{2}{*}{\texttt{activation}} & Params & FLOPs & TP. & Train TP. & Top-1 \\
    & (M) & (G) & (img/s) & (img/s) & acc. (\%) \\
    \hline
    SiLU & 30.2 & 4.91 & 1686 & 571 & 82.60 \\
    GELU & 30.2 & 4.91 & 1680 & 570 & 82.53 \\
    ReLU & 30.2 & 4.91 & 1684 & 577 & 82.65 \\
    \Xhline{1.0pt}
\end{tabular}
\end{table}

\newpage
\section*{NeurIPS Paper Checklist}

\begin{enumerate}

\item {\bf Claims}
    \item[] Question: Do the main claims made in the abstract and introduction accurately reflect the paper's contributions and scope?
    \item[] Answer: \answerYes{} 
    \item[] Justification: 
    The abstract and introduction provide a comprehensive overview of the background and motivation of this study, effectively outlining its main contributions point-by-point, thus accurately reflecting the paper's scope and significance.

\item {\bf Limitations}
    \item[] Question: Does the paper discuss the limitations of the work performed by the authors?
    \item[] Answer: \answerYes{} 
    \item[] Justification: 
        We primarily focused on discussing the limitations associated with this study in section~\ref{sec:conclusion_and_limitation}.

\item {\bf Theory Assumptions and Proofs}
    \item[] Question: For each theoretical result, does the paper provide the full set of assumptions and a complete (and correct) proof?
    \item[] Answer: \answerYes{} 
    \item[] Justification: 
    The paper includes the full set of assumptions and correct proofs for each theoretical result, primarily presented in the appendix. Notably, it covers the formulation of State Space Models, the discretization process, the derivation of recurrence relationships, and explanations concerning self-attention computations, ensuring completeness and accuracy in theoretical presentation.

\item {\bf Experimental Result Reproducibility}
    \item[] Question: Does the paper fully disclose all the information needed to reproduce the main experimental results of the paper to the extent that it affects the main claims and/or conclusions of the paper (regardless of whether the code and data are provided or not)?
    \item[] Answer: \answerYes{} 
    \item[] Justification: 
    All information regarding the key contribution of this paper, \textit{i.e.}, the introduction of selective SSMs (or S6) to processing vision data, as well as architectural and experimental configurations, have been fully disclosed (to the extent that it affects the main claims and/or conclusions of the paper). Furthermore, the implementation of other components within the proposed VMamba framework, such as Vision Transformers and the parallelized Selective Scan operation, is facilitated by the plenty of support available from existing open-source resources within the community.

\item {\bf Open access to data and code}
    \item[] Question: Does the paper provide open access to the data and code, with sufficient instructions to faithfully reproduce the main experimental results, as described in supplemental material?
    \item[] Answer: \answerYes{} 
    \item[] Justification: 
    The supplementary material submitted with the manuscript includes open access to all source code and scripts necessary to faithfully reproduce the main experimental results. Instructions for running the code are also provided within the scripts.

\item {\bf Experimental Setting/Details}
    \item[] Question: Does the paper specify all the training and test details (e.g., data splits, hyperparameters, how they were chosen, type of optimizer, etc.) necessary to understand the results?
    \item[] Answer: \answerYes{} 
    \item[] Justification: 
    The paper specifies detailed experimental configurations in Section~\ref{sec:appendix_details_for_vmamba_models} in Appendix, providing readers with essential information to comprehend the results. Following established conventions in the field of vision backbone models, the evaluation protocol encompasses standard practices commonly found in the relevant literature, ensuring readers can refer to established methodologies.

\item {\bf Experiment Statistical Significance}
    \item[] Question: Does the paper report error bars suitably and correctly defined or other appropriate information about the statistical significance of the experiments?
    \item[] Answer: \answerNo{} 
    \item[] Justification: 
    We did not include an analysis of the statistical significance of the experiments mainly due to the prohibitively expensive training cost of vision backbone models and our limited computing resources. However, we have provided the code, hyperparameters, and random seeds used in our experiments to facilitate the reproducibility of our findings. We would like to point out that, due to the extensive amount of training data, the statistical patterns of the experiment results are likely to remain consistent across different trials. Consequently, reporting error bars or other information about statistical significance is not a common practice in studies developing deep vision backbones.

\item {\bf Experiments Compute Resources}
    \item[] Question: For each experiment, does the paper provide sufficient information on the computer resources (type of compute workers, memory, time of execution) needed to reproduce the experiments?
    \item[] Answer: \answerYes{} 
    \item[] Justification: 
    All experiments were carried out on an 8 $\times$ A100 GPU server, as detailed at the beginning of the experiment section (Section~\ref{sec:experiments}).

\item {\bf Code Of Ethics}
    \item[] Question: Does the research conducted in the paper conform, in every respect, with the NeurIPS Code of Ethics \url{https://neurips.cc/public/EthicsGuidelines}?
    \item[] Answer: \answerYes{} 
    \item[] Justification: 
    After carefully reviewing the referenced document, we certify that the research conducted in the paper conforms, in every respect, with the NeurIPS Code of Ethics.

\item {\bf Broader Impacts}
    \item[] Question: Does the paper discuss both potential positive societal impacts and negative societal impacts of the work performed?
    \item[] Answer: \answerNA{} 
    \item[] Justification: 
        The paper primarily focuses on vision backbones trained using publicly available datasets that have undergone thorough validation. While the vision backbone itself is not directly applicable to everyday scenarios, it serves as a neutral and valuable toolkit for further development and research.
    
\item {\bf Safeguards}
    \item[] Question: Does the paper describe safeguards that have been put in place for responsible release of data or models that have a high risk for misuse (e.g., pretrained language models, image generators, or scraped datasets)?
    \item[] Answer: \answerNA{} 
    \item[] Justification: 
        The proposed models are vision backbone networks trained on benchmark datasets such as ImageNet-1K, MSCOCO, and ADE20K. These datasets have been extensively used in the computer vision community and have undergone comprehensive safety risk assessments.

\item {\bf Licenses for existing assets}
    \item[] Question: Are the creators or original owners of assets (e.g., code, data, models), used in the paper, properly credited and are the license and terms of use explicitly mentioned and properly respected?
    \item[] Answer: \answerYes{} 
    \item[] Justification: 
    In the paper, we specified the datasets and code sources used (e.g., mmdet), and provided appropriate citations in the reference section. Additionally, we ensured transparency by including the sources of any modified code files, making the changes traceable.

\item {\bf New Assets}
    \item[] Question: Are new assets introduced in the paper well documented and is the documentation provided alongside the assets?
    \item[] Answer: \answerYes{} 
    \item[] Justification: 
    We have included the code, along with detailed usage instructions, in the supplementary materials. After the review process is completed, we will make the code publicly available to the community.

\item {\bf Crowdsourcing and Research with Human Subjects}
    \item[] Question: For crowdsourcing experiments and research with human subjects, does the paper include the full text of instructions given to participants and screenshots, if applicable, as well as details about compensation (if any)? 
    \item[] Answer: \answerNA{} 
    \item[] Justification: 
        This study does not involve any crowdsourcing experiments or research with human subjects.

\item {\bf Institutional Review Board (IRB) Approvals or Equivalent for Research with Human Subjects}
    \item[] Question: Does the paper describe potential risks incurred by study participants, whether such risks were disclosed to the subjects, and whether Institutional Review Board (IRB) approvals (or an equivalent approval/review based on the requirements of your country or institution) were obtained?
    \item[] Answer: \answerNA{} 
    \item[] Justification: 
        No crowdsourcing experiments or research with human subjects were involved in this study. All experiments were conducted using code and GPU servers.

\end{enumerate}

\end{document}